\documentclass[conference]{IEEEtran}
\IEEEoverridecommandlockouts
\usepackage{amsmath,amssymb,amsfonts}
\usepackage[numbers, sort]{natbib}
\usepackage{algorithmic}
\usepackage{graphicx}
\usepackage{textcomp}
\usepackage{xcolor}
\def\BibTeX{{\rm B\kern-.05em{\sc i\kern-.025em b}\kern-.08em
    T\kern-.1667em\lower.7ex\hbox{E}\kern-.125emX}}

\usepackage{tikz}
\usepackage{balance}  % for  \balance command ON LAST PAGE  (only there!)
\usepackage[linesnumbered,ruled,noend]{algorithm2e}
\usepackage{subfigure}
\usepackage{booktabs}
\usepackage{enumitem}
\usepackage{amsthm}

\usepackage{multirow}
\usepackage{listings}
\usepackage{balance}

\setlist[itemize]{leftmargin=*}

\newcommand{\ourmethods}[1]{EvoMoE}

\newcommand{\moe}{MoE}
\newcommand{\evomoe}{{EvoMoE}\xspace}

\begin{document}

\title{EvoMoE: An Evolutional Mixture-of-Experts Training Framework via Dense-To-Sparse Gate}

\author{
\centering
\IEEEauthorblockN{Xiaonan Nie, Xupeng Miao, Zhi Yang, Bin Cui}
\IEEEauthorblockA{
Department of Computer Science, Peking University, Beijing, China \\
\{xiaonan.nie, xupeng.miao, yangzhi, bin.cui\}@pku.edu.cn}
}
\author{\IEEEauthorblockN{
Xiaonan Nie$^{\dag}$~~~~~Xupeng Miao$^{\dag\$}$~~~~~Shijie Cao$^\ddagger$~~~~~Lingxiao Ma$^\ddagger$~~~~~Qibin  Liu$^\dag$~~~~~Jilong Xue$^\ddagger$ \\ Youshan Miao$^\ddagger$~~~~~~Yi Liu$^\#$~~~~~~Zhi Yang$^{\dag}$~~~~~Bin Cui$^{\dag\S}$}
\IEEEauthorblockA{\textit{$^{\dag}$School of Computer Science \& Key Laboratory of High Confidence Software Technologies (MOE), Peking University} \\
\textit{$^\$$Carnegie Mellon University}\quad 
\textit{$^\ddagger$Microsoft Research}\quad \textit{$^\#$Tencent Inc}\\
\textit{$^\S$Institute of Computational Social Science, Peking University (Qingdao), China}\\
\textit{\{xiaonan.nie, xupeng.miao, 1700012767, yangzhi, bin.cui\}@pku.edu.cn}\\
\textit{\{lingxiao.ma, shijiecao, jxue, yomia\}@microsoft.com}~~~~~~\textit{callbackliu@tencent.com}}
}

\maketitle

\begin{abstract}
Mixture-of-experts (MoE) is becoming popular due to its success in improving the model quality, especially in Transformers.
By routing tokens with a sparse gate to a few experts (i.e., a small pieces of the full model), MoE can easily increase the model parameters to a very large scale while keeping the computation cost in a constant level.
Most existing works just initialize some random experts, set a fixed gating strategy (e.g., Top-$k$), and train the model from scratch in an ad-hoc way. We identify that these MoE models are suffering from the immature experts and unstable sparse gate, which are harmful to the convergence performance.

In this paper, we propose an efficient end-to-end MoE training framework called \evomoe. \evomoe starts from training one single expert and gradually evolves into a large and sparse MoE structure.
\evomoe mainly contains two phases: the \textit{expert-diversify} phase to train the base expert for a while and spawn multiple diverse experts from it,
and the \textit{gate-sparsify} phase to learn an adaptive sparse gate and activate a dynamic number of experts.
\evomoe{} naturally decouples the joint learning of both the experts and the sparse gate and focuses on learning the basic knowledge with a single expert at the early training stage. Then it diversifies the experts and continues to train the MoE with a novel Dense-to-Sparse gate (DTS-Gate). Specifically, instead of using a permanent sparse gate, DTS-Gate begins as a dense gate that routes tokens to all experts, then gradually and adaptively becomes sparser while routes to fewer experts.
Evaluations are conducted on three popular models and tasks, including RoBERTa for masked language modeling task, GPT for language modeling task and Transformer for machine translation task. 
The results show that \evomoe{} outperforms existing baselines, including Switch, BASE Layer, Hash Layer and StableMoE. Specifically, \ourmethods{} outperforms other MoE methods on GLUE benchmark up to 0.562 and 0.403 on average. 
Our code is available \footnote{https://github.com/codecaution/EvoMoE}.
\end{abstract}


%
%-------------------------------------------------------------------------------
\section{Introduction}
%-------------------------------------------------------------------------------
\label{sec:intro}

\begin{figure*}[t]
    \centering
    \includegraphics[width=0.8\textwidth]{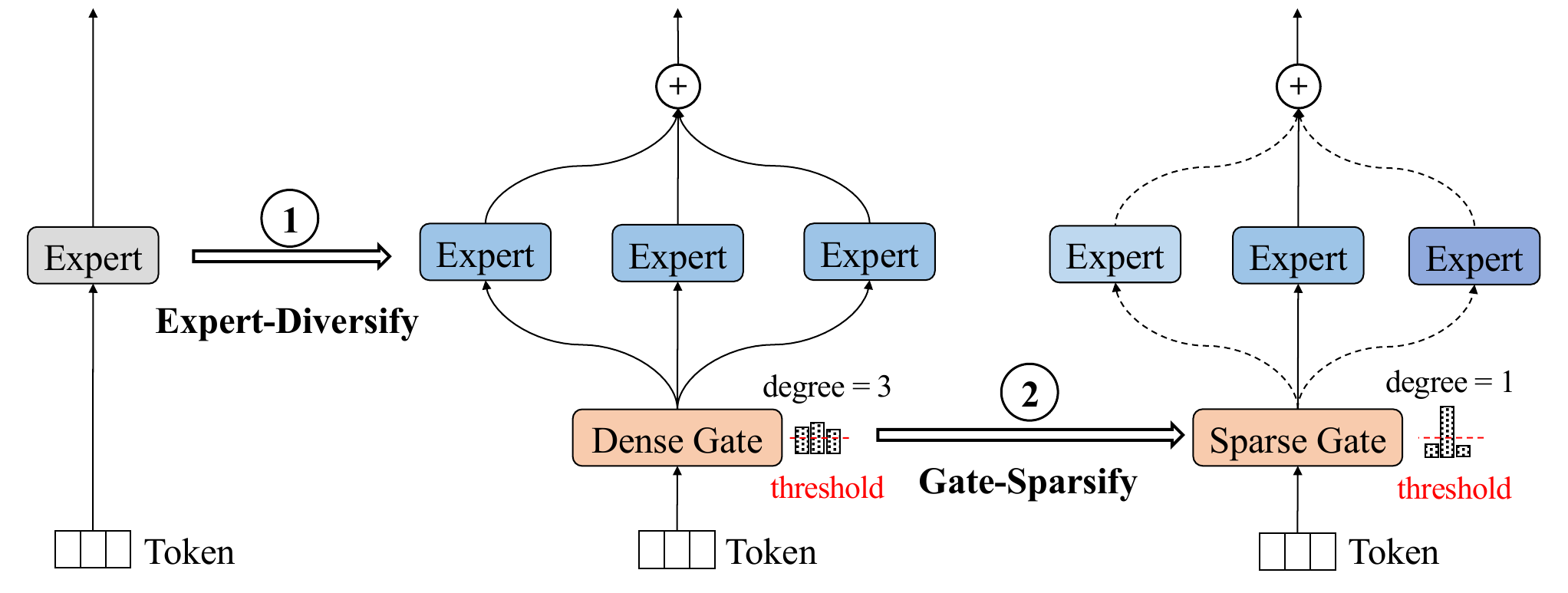}
    \caption{Illustration on the workflow of \ourmethods{}, which contains two phases: (1) expert-diversify phase and (2) gate-sparsify phase. In the first stage, we train one shared-expert instead of $N$ individual experts and then adopts $diversify$ functions (i.e., random masking) to spawn multiple diverse experts from the shared expert. In the second stage, we propose the Dense-to-Sparse gate, which starts as a dense gate that routes tokens to most experts and then gradually becomes sparser. Different from previous TopK-based gates, we propose the content-based gating mechanism, which activates experts whose weight is beyond the threshold.} 
    \label{fig:overview}
\end{figure*}

% introduce transformer and moe
The Transformer model architecture is becoming increasingly important in data mining and has achieved impressive results in a wide range of applications, such as natural language processing~\cite{DBLP:conf/nips/gpt3}, computer vision~\cite{DBLP:ViT}, graph learning~\cite{graphormer}, and recommendation systems~\cite{chen2019behavior}. Recently, there is a trend of improving the capability of Transformer models through enlarging data and model scales \citep{DBLP:conf/nips/gpt3}. Specifically, \cite{DBLP:scaling_laws} explored the scaling law of transformer models which shows that the model performance scales as a power-law with data sizes, model sizes and the computation. However, with the rapid increasing of the model sizes, it is hard to further scale the model to extremely large sizes due to the limited computation power of available hardware devices.
To address these challenges, sparsely-gated Mixture-of-Experts (MoE), a popular form of conditional computation, has been proposed to increase the model size while without increasing the computational cost (e.g., FLOPs) proportionally~\citep{DBLP:journals/corr/BengioLC13, DBLP:lstm_moe, DBLP:switch, DBLP:gshard, lewis2021base, roller2021hash}. 
Specifically, the input tokens are routed by a sparse gate to a few experts, leading to lower computational costs compared to a dense model with the same model size.

The success of MoE model relies on both the large model capacity introduced by plenty of experts and the sophisticated sparse routing connections learned by the gate network. 
Many existing works~\citep{DBLP:switch, DBLP:gshard, lewis2021base, roller2021hash} are exploring novel gating networks to improve the model quality or the training efficiency.
They typically adopt a pre-defined sparse gate architecture (e.g., Top-1 or Top-2 with a fixed number of activated experts), and then train the model parameters of both the gate and experts jointly from scratch.
However, such \textit{joint-training over pre-defined sparse architecture} could severely limit the model quality, and even the training efficiency.
Particularly, at the beginning of training a MoE model, both the gate and the experts are randomly initialized. The gate does not have evidence to decide which expert to process an input token, and the experts also do not have experiences to process a randomly-assigned input token. Training all experts from a random state with random routed samples requires a long and duplicated warming-up process.
Furthermore, these pre-defined gates limit the MoE to explore only 1 or 2 experts at a time. But in the early stage with a immature gate, such small opportunities could be easily  influenced by the random routing noises, and the improper routing could even be reinforced for a long time.
Our observation shows that such random routing in the initial stage and long-distance reinforce-based model updating in existing approaches could affect both the training time and final model quality.

In this paper, to overcome the limitations in existing approaches, we revisit the learning process of MoE models and advocate a simple but effective end-to-end training paradigm, named \evomoe{}. 
Instead of directly training from a pre-defined sparse architecture, \evomoe{} gradually evolves into a diverse and sparse MoE architecture from an initial model in two phases: \textbf{expert-diversify} and \textbf{gate-sparsify}.
Specifically, we find that both the gate and the experts are underachieving in MoE training thus resulting in unstable routing performance. Motivated by the successes of weight-sharing models, in the first stage, we introduce the expert-diversify mechanism to learn the commonly shared knowledge across different experts.
Our proposed mechanism only trains one common experts with all input tokens at first, which could be seen as sharing the model parameters across all the experts.
To involve the diversity of these experts, we then randomly perturb each expert with different masks as the initial model state of the following training steps.
In the gate-sparsify phase, the weight-sharing constraint is released and the training of MoE turns to the sparsely activated manner over these diverse experts.
Unlike the pre-defined sparse gate in the previous works, we introduce DTS (Dense-to-Sparse) gate to decide the sparse gate gradually for MoE models. We proposed DTS gate to adaptively learn a better gating network from a dense one and gradually route tokens to fewer experts, making the training structure sparser and continuously reducing the computation cost, while keeping the model quality improving as usual.
In particular, to implement the DTS gate, our idea is to carefully control the temperature of a softmax-based routing function, so that to adjust the weights distribution among experts and control the sparsity of the MoE layer during training.

In short, \evomoe advances in two aspects. First, compared to the \textit{joint-training of gate and experts} from scratch, \evomoe splits joint-training process and provides an opportunity to train experts during a warm start. Such mechanism of \textit{training gate after experts} can reduce a lot of random error-trails at the beginning. Second, compared to the reinforce-based model updating, starting with a dense gate allows us to get training feedback from all diverse experts and adjust the routing weights directly to the right direction, which not only speeds up the convergence of the gate, but also benefits for the expert specialization.

We evalute \evomoe{} on three popular models and tasks, including RoBERTa~\cite{liu2019roberta} (Encoder-Only) for masked language modeling (MLM) task, GPT~\cite{gpt2} (Decoder-Only) for language modeling (LM) task and Transformer~\cite{DBLP:conf/nips/VaswaniSPUJGKP17} (Encoder-Decoder) for machine translation (MT) task. 
The results show that \evomoe{} outperforms existing baselines, including Switch~\cite{DBLP:switch}, BASE Layer~\cite{DBLP:conf/icml/baselayer}, Hash Layer~\cite{roller2021hash} and StableMoE~\cite{DBLP:conf/acl/stablemoe}. Specifically, on MLM task, \ourmethods{} outperforms other MoE methods up to 0.562 GLUE score and 0.403 in average for the GLUE benchmark~\cite{DBLP:conf/emnlp/glue}; 
on LM task,  \ourmethods{} outperforms other MoE methods up to 0.88 ppl and 0.545 ppl on average;
on translation task, \evomoe{} can averagely achieve 1.0 BLEU score improvement as well as 1.33x speed-up that Switch Transformer.
Experiments also verify the ability of \evomoe for scaling models with more experts or more MoE layers.

The rest of the paper is organized as follows. We first introduce the background of Transformers and MoEs in Section~\ref{sec:preliminary}. And we identify two key defects in existing MoE training process including the conformity and instability in Section~\ref{sec:motivation}. Motivated by these properties, we present our \evomoe design in Section~\ref{sec:methods} and introduce the expert-diversity stage and gate-sparsify stage respectively. Section~\ref{sec:impl} describes some implementation details.
% list the relevant existing methods in Section~\ref{sec:related_work}. Then, in , we represent the xxxx.
We provide the evaluation methodology and conduct substantial experiments under various settings in section~\ref{sec:experiment} to support our claims. More relevant approaches are introduced in Section~\ref{sec:related_work}. Finally, we provide some concluding remarks in section~\ref{sec:conclusion}.

\begin{figure}[t]
    \centering
    \subfigure[Transformer Layer]{
            \includegraphics[width=0.13\textwidth]{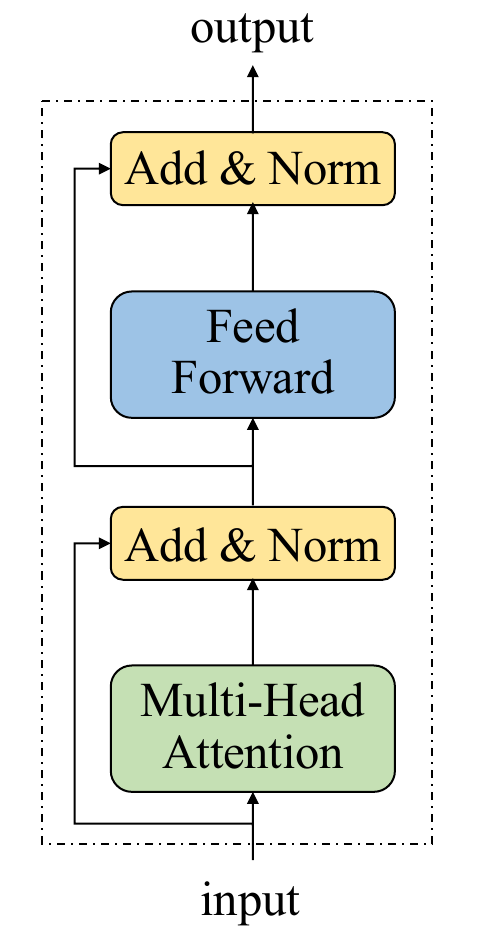}
            \label{fig:transformer}
    }
    \subfigure[Transformer Layer with Mixture-of-Expert]{
            \includegraphics[width=0.32\textwidth]{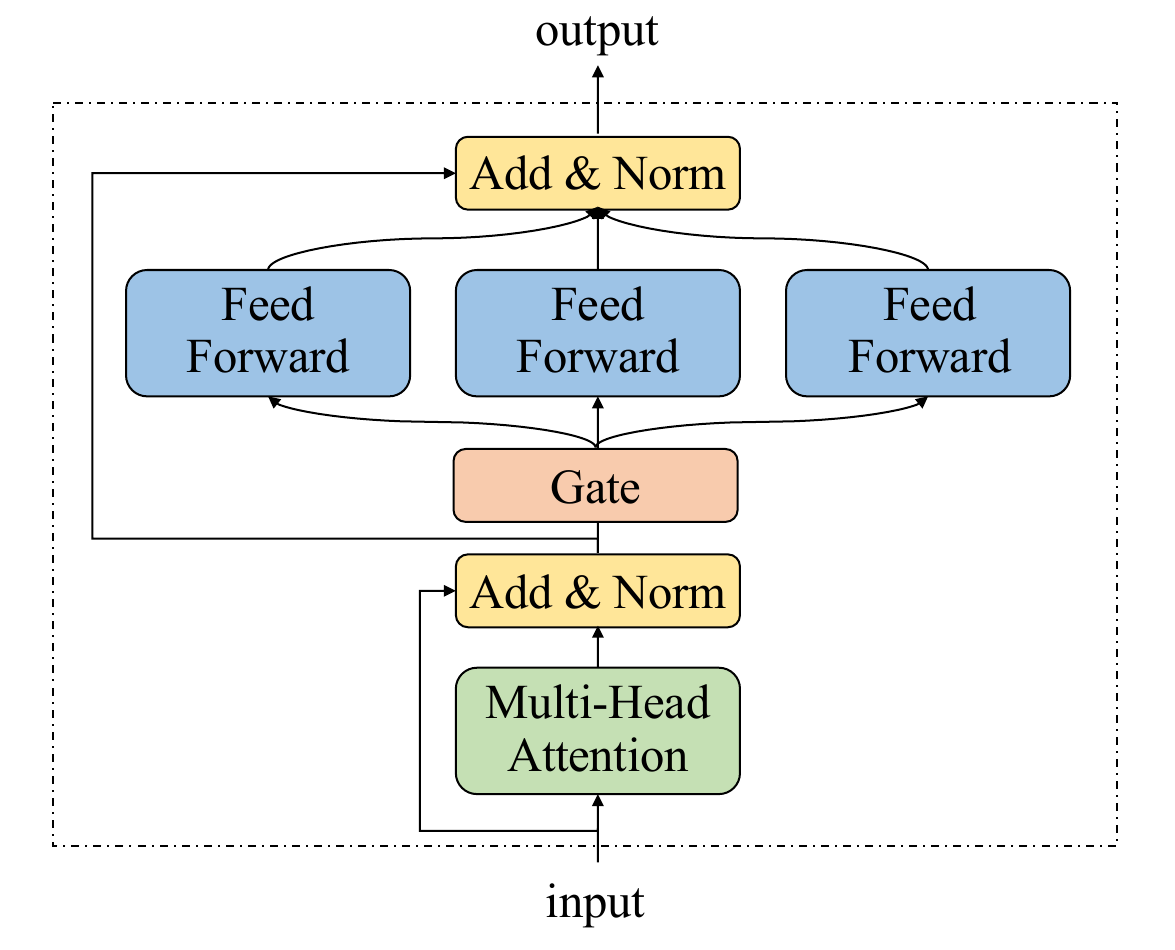}
            \label{fig:moe_transformer}
    }
    \caption{A brief architecture of Transformer Encoder Layer and Transformer with Mixture-of-Expert Layer. The Transformer encoder layer contains two main components: a Multi-Head Self-Attention Layer and a Position-wise Feed-Forward Layer. Based on Transformer layer, the transformer with MoE replaces the FFN with a series of FFNs and introduce a gate network.}
    \label{fig:preliminary}
\end{figure}

\section{Preliminary}
\label{sec:preliminary}
\subsection{Transformer}

The model architecture of Transformer~\cite{vaswani2017attention} has demonstrated its superior performance in many sequence-to-sequence natural language processing (NLP) tasks, which contains several encoder layers and decoder layers. Each encoder layer is stacked by a multi-head self-attention network and a position-wise feed-forward network (FFN), which is illustrated in Figure~\ref{fig:transformer}. 
It employs a residual connection on each of these two sub-layers, followed by a normalization layer~\cite{DBLP:journals/corr/layernorm}. Formally, each sub-layer, e.g., attention and FFN, produces its output as $\text{LayerNorm}(x + \text{Sublayer}(x))$.
The decoder is similarly constructed, except for an additional cross-attention mechanism between attention and FFN to introduce the output of the encoder. For a sequence of input tokens $(x_{1},...,x_{s})$ $\in$ $\mathbb{R}^{D}$, we formulate the function of each sub-layer in following:

\begin{table}[t]
\centering
\renewcommand{\multirowsetup}{\centering}
\caption{Notations}
\label{tab:notation}
\vspace{-2mm}
\begin{tabular}{cc}
\toprule 
Symbols & Definitions \\
\midrule
$Q$ & Queries in self-attention module \\
\midrule
$K$ & Keys in self-attention module \\
\midrule
$V$ & Values in self-attention module\\
\midrule
$d_k$ & Feature dimension of each Query/Key \\
\midrule
$S$ & A set of input tokens \\
\midrule
$E$ & An series of experts\\
\midrule
$D$ & Feature dimension of each token \\
\midrule
$N$ & Number of experts in each MoE layer \\
\midrule
$e_{i}(x_s)$ & The output of $i$-$th$ Expert by taking input $x_s$ \\
\midrule
$g(x_s)_i$ & The routing score for $x_s$ on $i$-$th$ Expert\\
\midrule
$c$ & Threshold for expert selection \\

\midrule
$G_{S, E}$ & The routing score for $S$ on $E$ \\
\midrule
$Id_{S}$ & The set of selected expert id on $S$\\
\bottomrule
\end{tabular}
\end{table}

\textbf{Attention:} The attention module~\cite{DBLP:conf/nips/VaswaniSPUJGKP17} could capture the dependencies between tokens in the sequence, and is effective in sequence modeling. It performs as a ternary function, which maps the input queries (Q), keys (K) and values (V) to the output (O). Equation~\ref{equ:attention} represents the Scaled Dot-Product Attention~\cite{DBLP:conf/nips/VaswaniSPUJGKP17}, which performs dot products of each query with all keys, divides each by $\sqrt{d_k}$ and then adopts the softmax function to get the weight of each value. In addition, $d_k$ is the dimension of queries and keys.
\begin{equation}
    \texttt{Attention}(Q, K, V) = \texttt{softmax}(\frac{QK^T}{\sqrt{d_k}})V
\label{equ:attention}
\end{equation}

\textbf{Multi-Head Attention:} ~\citet{DBLP:conf/nips/VaswaniSPUJGKP17} proposed the multi-head attention mechanism to jointly learn from different representation subspaces at different positions and thus improved the model performance. The multi-head attention linearly projected the queries, keys and values $h$ times with learned linear projections to $d_k$, $d_k$ and $d_v$, dimensions, respectively. 

\begin{align}
    \texttt{MultiHead}&(Q, K, V) = \texttt{Concat}(head_1, ..., head_h){W^O} \\
    \text{where}&\ head_i = \texttt{Attention}(QW_{i}^Q, KW_{i}^K, VW_{i}^V) \nonumber
\label{equ:multi_head}  
\end{align}

\noindent The projections are the trainable parameter matrices, where $W_{i}^Q \in \mathbb{R}^{d_{model} \times d_k}$, $W_{i}^K \in \mathbb{R}^{d_{model} \times d_k}$, $W_{i}^V \in \mathbb{R}^{d_{model} \times d_v}$. Meanwhile, $h$ is the number of heads, and $d_k = d_v = d_{model}/h$. Because the dimension of each head is reduced from $d_{model}$ to $d_{model}/h$, the time cost of multi-head attention is similar to that of the original attention. In addition, the decoder employs a masked self-attention, which only sees the tokens on the left of the sequence.

\textbf{Position-wise Feed-Forward Networks:} Each transformer layer also includes a fully connected feed-forward network (Equation~\ref{equ:ffn}), which consists of two fully connected networks and a ReLU activation function.

\begin{equation}
    \texttt{FFN}(x_{s}) = W_{2} \cdot \texttt{ReLU}(W_{1} \cdot x_{s} + b_{1}) + b_{2}
\label{equ:ffn}
\end{equation}

\begin{figure}[t]
    \centering
    \includegraphics[width=0.415\textwidth]{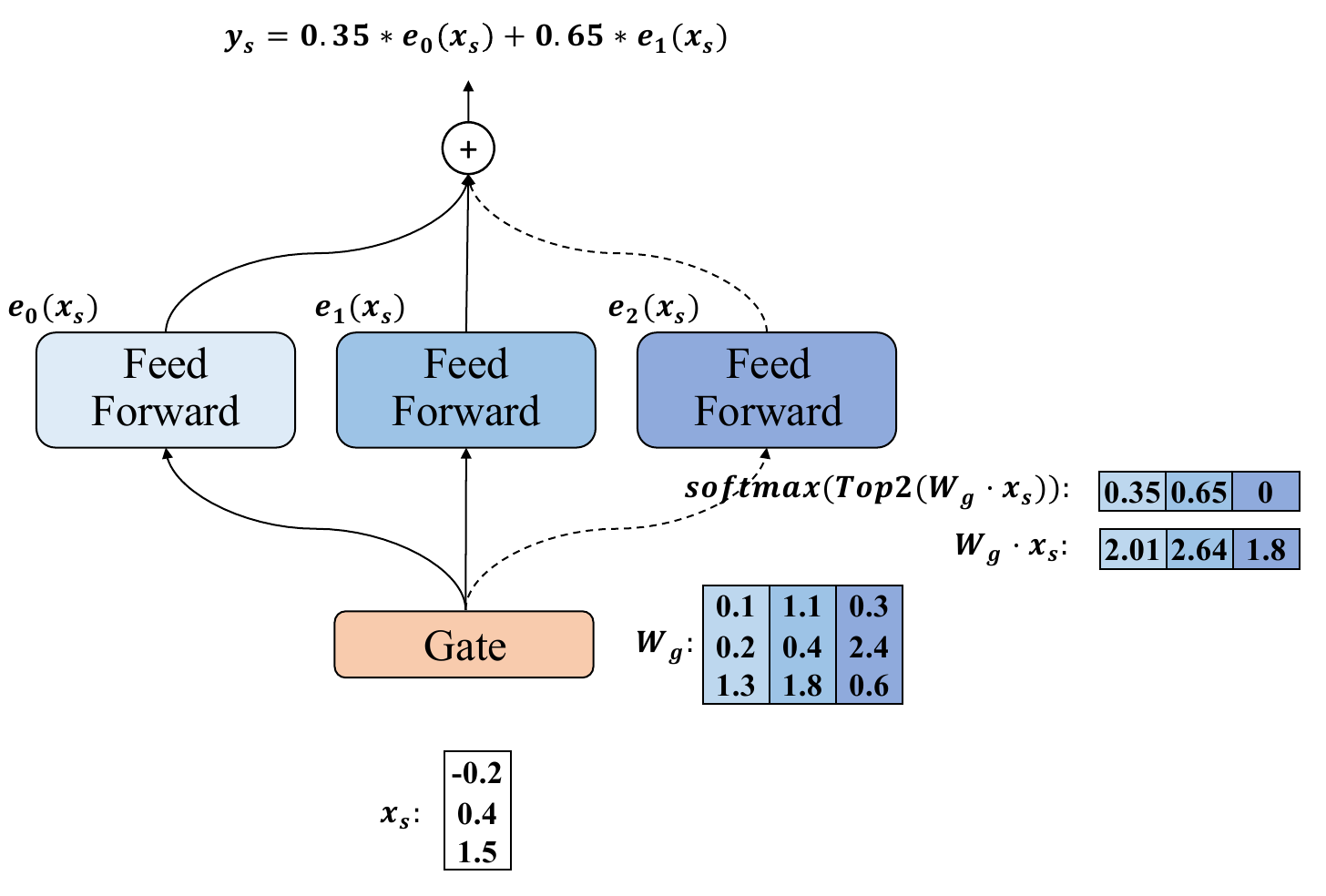}
    \caption{Illustration on the workflow of Mixture-of-Expert (MoE). The input token $x_s$ is first processed by the gate network to calculate similarities between $x_s$ and each expert. Then, it performs Top-K operation on these similarities to determine the target activated experts. Finally, $y_s$ is produced by the linearly weighted combination of each expert's output on the token by the gate's output. }
    \label{fig:moe_layer}
\end{figure}

\begin{figure*}[t]
    \centering
    \subfigure[Expert Loads Distribution .]{
            \includegraphics[width=0.22\textwidth]{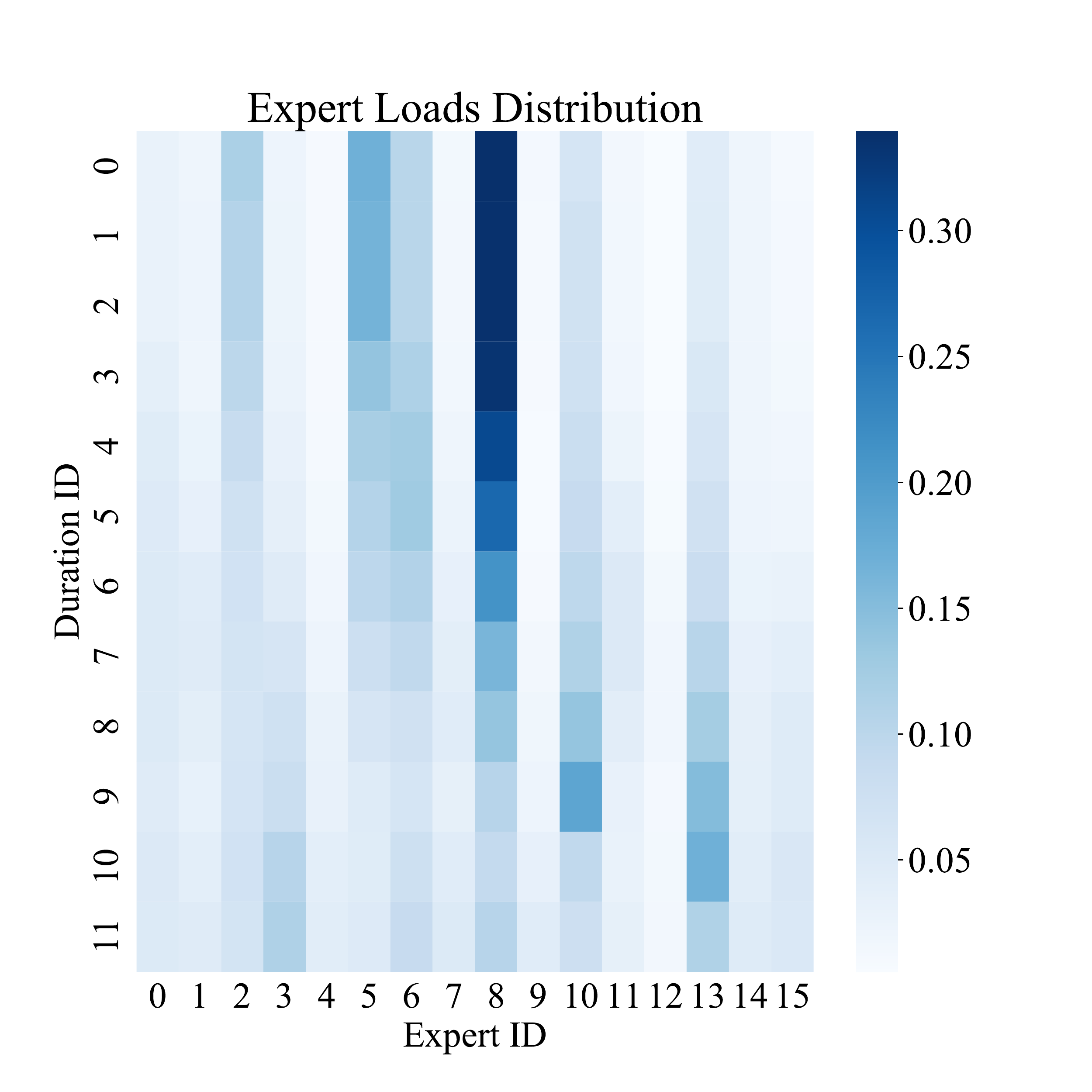}
            \label{fig:duration_distribution}
    }
    \subfigure[Unstable Routing Pattern for Token ``the'']{
            \includegraphics[width=0.68\textwidth]{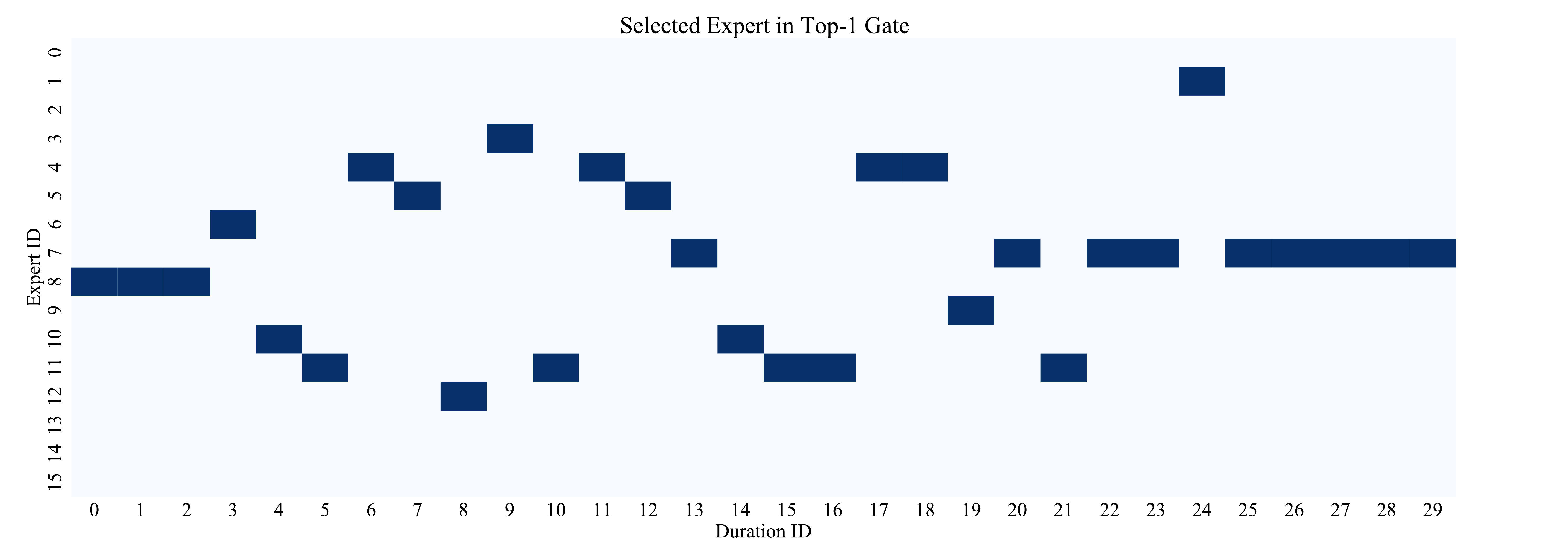}
            \label{fig:weights_distribution}
    }
    \caption{The observations of GPT-MoE with 16 experts and the Top-1 gate among 12 MoE-layers (totally 24 layer).
    Figure~\ref{fig:duration_distribution} shows the expert load distribution (deeper color represents heavier workload, i.e., more tokens to be processed) and Figure~\ref{fig:weights_distribution} shows the expert selection for a specific token ``the''.}
    \label{fig:dense_moe}
\end{figure*}

\subsection{Mixture of Experts}
Because larger pretrained models always achieve better model quality~\cite{DBLP:scaling_laws}, the size of state-of-the-art NLP models has been increasing $10\times$ per year, e.g., BERT~\cite{DBLP:conf/naacl/BERT}, GPT~\cite{gpt}, T5~\cite{DBLP:googleT5}, GPT-2~\cite{gpt2}, GPT-3~\cite{DBLP:conf/nips/gpt3}, which require increasing compute budgets. To improve the model capacity without increasing computation budgets, researchers sparsely scale transformers recently as Figure~\ref{fig:moe_transformer} by replacing the feed-forward network with the mixture of experts (MoE) architecture and activating only a subset of these experts for each input sample~\cite{DBLP:lstm_moe, DBLP:switch, DBLP:gshard}.
The main components of the MoE architecture include an expert network $E$ for scaling model capacity and a sparse gate network $G$ for introducing model sparsity.

\textbf{Expert Network:} 
The expert network $E$ includes a series of experts $\{e_{1},...,e_{N}\}$ to increase the model capacity, where each expert $e_{i}$ represents a single neural network, e.g., \text{FFN}, and contains its own parameters. In Figure~\ref{fig:moe_layer}, the MoE layer consists of three FFN networks. 
For each expert $e_{i}$ ($e_{i}: \mathbb{R}^{D} \rightarrow \mathbb{R}^{D}$), it takes the token $x_{s}$ as an input to produce its own output $e_{i}(x_{s})$. The final output of the expert network $y_{s}$ is the linearly weighted combination of each expert’s output on the token by the gate's output, formulated as Equation \ref{equ:moe_weighted_sum}. 
\begin{equation}
    y_{s} = \sum_{i=1}^{N}g(x_{s})_{i} \cdot e_{i}(x_{s})
\label{equ:moe_weighted_sum}
\end{equation}
\noindent In Figure~\ref{fig:moe_layer}, the expert network takes the input token $x_{s}: [-0.2, 0.4, 1.5]$ and produces the output of each individual expert on $x_s$, e.g., $e_{0}(x_s)$, $e_{1}(x_s)$ and $e_{2}(x_s)$. By combining the gate's output, i.e., $[0.35, 0.65, 0]$, the output of this MoE layer is $y_s = 0.35*e_{0}(x_s) + 0.65*e_{1}(x_s)$.

\textbf{Sparse Gate Network:}
The sparse gate network $G$ is the key component to introduce model sparsity, which takes a batch of tokens $\{x_{1},...,x_{s}\}$ as input and produces the probability of them with respective to all experts $\{e_{1},...,e_{N}\}$.
\citet{DBLP:lstm_moe} proposes the Top-K gating as Equation~\ref{equ:topk_gate}, which keeps only the top k values before the softmax function. In addition, $W_{g}$ is a trainable variable ($W_{g} \in \mathbb{R}^{D \times N}$) and determine the targeted experts for each token.

% Implement topK before or after the softmax is not certain for these works.

\begin{equation}
    g(x_s) = softmax(TopK(x_{s} \cdot W_{g}) )
\label{equ:topk_gate}
\end{equation}
\noindent We illustrate the workflow of a MoE layer in Figure~\ref{fig:moe_layer}, where $k=2$ and $W_g$ is a 3 $\times$ 3 (i.e., feature dimension $\times$ number of experts) matrix to represents the parameter of gate network. We first perform a dot-product on $x_s$ and $W_g$ to calculate similarity between the input token and the experts. The result, $[2.01, 2.64, 1.8]$, indicates that the input prefers $e_1 > e_0 > e_2$ and we only activate $e_0$ and $e_1$ as $k = 2$. Finally, we conduct a softmax function to get the weight score of each expert and perform a weighted sum to get the final output $y_s$. 

Previous work mainly focuses on how to improve the quality and efficiency of training such sparse gate network. ~\citet{DBLP:lstm_moe} proposed the noisy Top-K gating on Long Short-TerM memory (LSTM) kayers~\cite{DBLP:journals/neco/lstm} and ~\citet{DBLP:gshard} introduced MoE with Top-2 gate into Transformer.
~\citet{lewis2021base} adopted the numerous solution for balanced  token-to-expert routing and \citet{roller2021hash} utilized the hash-based routing strategy.

\textbf{Distributed Training of MoE Models:}
Expert parallel training is a specific method of parallelism for MoE models, which is first proposed by GShard~\cite{DBLP:gshard}. Experts are placed on different workers and each worker takes a different batch of training samples. For non-MoE layers, expert parallelism behaves the same as data parallelism. In MoE layers, tokens in the sequence are sent to workers where their desired experts reside. Similar to model parallelism, the outputs of each MoE layer are exchanged again to be organized back into original sequences for the computation of the next layer. As MoE models often have numerous experts, expert parallelism can scale up with model size better than model parallelism.

\subsection{Observation and Motivation}
\label{sec:motivation}
In this section, we revisit the learning process of MoE models and introduce our two key findings in the following, which motivates us to design our \evomoe framework.

\textbf{Conformity in Mixture of Experts:} One interesting finding is conformity. During the early training stage, existing join-training methods of sparse MoE make the routing decision to comply with most tokens.
Here we train a GPT model including 24 transformer layers, with every FFN layer replaced by 16-expert MoE layer using the Top-1 gate. 

Figure~\ref{fig:duration_distribution} shows that most tokens keep concentrating on the 8-th expert at first, since it has been greedily reinforced.
After around hundreds of training steps (i.e., 1 duration equals 40 training steps), the other experts gradually catch up and the workload becomes balanced. Such phenomenon motivates us to focus on training a common expert and utilize the computational resources to accelerate the early stage.

\textbf{Instability in Mixture of Experts:} Another important finding is the instability. We take a single token ``the'' as an example and Figure~\ref{fig:weights_distribution} shows its expert selection results for a longer training process. As we can see, the selection is highly unstable later since both the gate network and the experts are not knowledgeable enough to obtain a stable routing pattern, especially at the early stage of the training process. This indicates that a pre-defined gates (e.g., Top-K) in existing works, which assumes a fixed number of activated experts, could limit the exploration of potential valuable experts. Aggressively increasing the number of activated experts could improve the model capacity but inherently violates the original design intention of sparse MoE. Such a dilemma motivates us to design an adaptive solution to balance the trade-off between the convergence performance and computation costs.

\begin{figure}[t]
    \centering
    \includegraphics[width=0.25\textwidth]{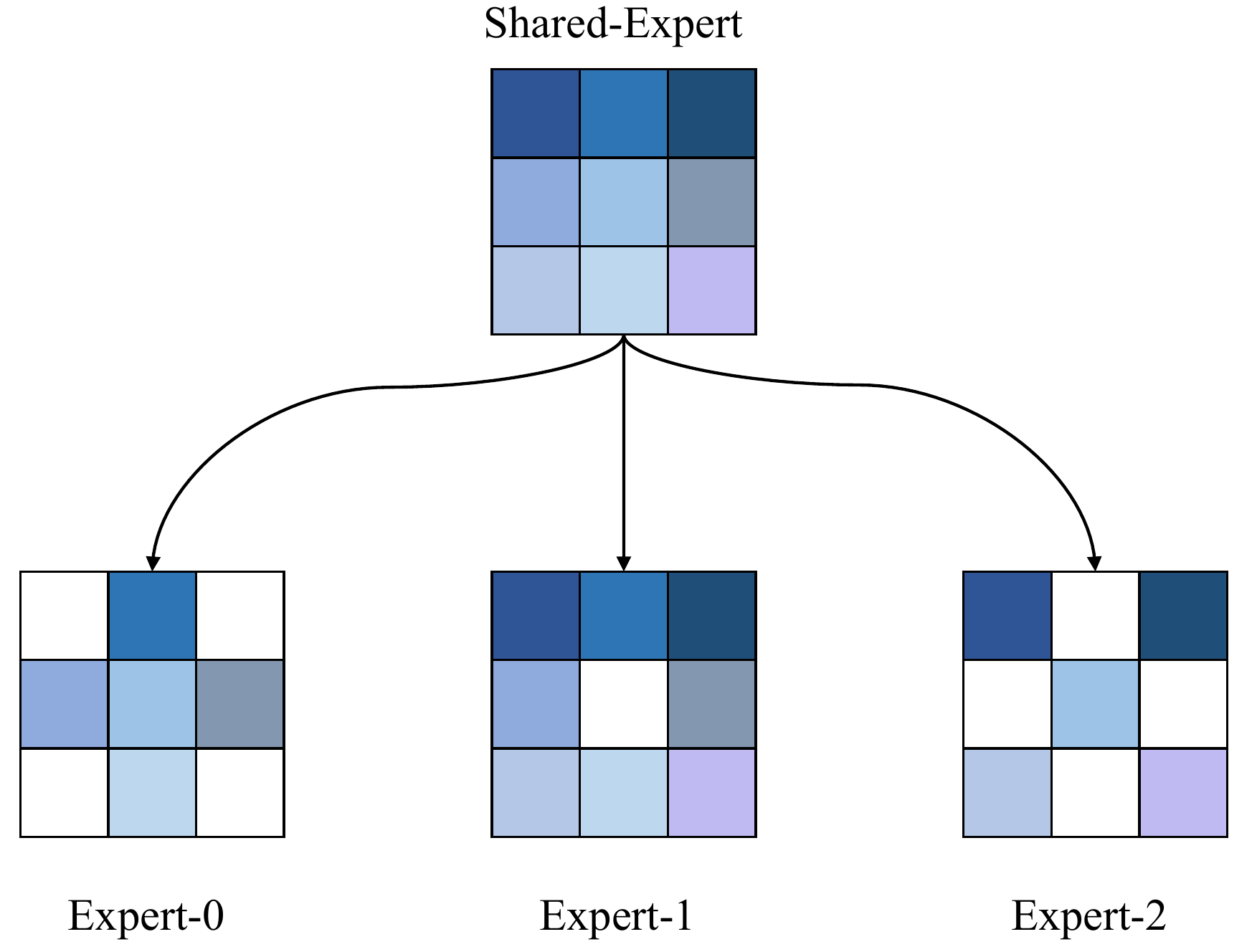}
    \caption{
    To spawn multiple diverse experts from the shared-expert, \ourmethods{} adopt the random masking technique. Specifically, 
    part of the shared expert's weight are masked as 0.}
    \label{fig:shared_to_diverse}
\end{figure}

\section{Methods}
\label{sec:methods}
The observations in Section~\ref{sec:motivation} motivates EvoMoE, a two-phase framework that gradually and adaptively training MoE-based models, which is different from existing methods that jointly train the gate network and the expert network over a pre-defined sparse (e.g., Top-1 or Top-2) gate and a series of randomly initialized experts.
As shown in Figure~\ref{fig:overview}, EvoMoE contains two phases: an expert-diversify phase and a gate-sparsify phase.
In the expert-diversify phase, \ourmethods{} shares the weights among experts in one MoE layer for several training steps and then makes experts diverse by randomly masking.  
In the gate-sparsify phase, \ourmethods{} introduces the dense-to-sparse (i.e., DTS) gate, which begins routing as a dense gate that routes tokens to all experts and then adaptively learns the weights of routing to each expert and gradually anneals to standard Top-1 gating.

\subsection{Problem Formulation}
Given an input token $x_s$, a series of experts $\{e_1,..., e_N\}$ and a learn-able gate with parameter $W_g$, \texttt{Func} is adopted by the gate network to determine the targeted experts for it, i.e., the token-to-expert assignment, formulated in Equation~\ref{equ:gate_problem}. $g(x_s)$ is a 1$\times$N vector, which represents the scores of $x_s$ with respect to experts.
Meanwhile, each expert will process the input token separately as $e_i(x_s)$ and combine their output as Equation~\ref{equ:moe_weighted_sum_problem}.

\begin{equation}
    g(x_s) = \texttt{Func}(x_{s} \cdot W_{g})
\label{equ:gate_problem}
\end{equation}

\begin{equation}
    y_{s} = \sum_{i=1}^{N}g(x_{s})_{i} \cdot e_{i}(x_{s})
\label{equ:moe_weighted_sum_problem}
\end{equation}

Existing work adopts a pre-defined Top-K as $Func$, such as Top-2 for GShard ~\cite{DBLP:gshard} and Top-1 for Switch-Transformer~\cite{DBLP:switch}. However, due to the non-derivability of Top-K, only the selected experts would back-propagate their gradients to the gate network and update their corresponding columns in $W_g$.
For example, only 1 expert is selected and 1 column of the gate would be updated in Switch-Transformer. So it is hard for Top-K gate to optimize this \textbf{expert-selection} problem.
Moreover, as observed in Figure~\ref{fig:duration_distribution}, the loads of experts are extremely imbalanced at the early stage of training and thus most GPUs suffer from low utilization due to stragglers in expert parallelism.

\setlength{\textfloatsep}{0.1cm}
\begin{algorithm}[t]
	\SetAlgoLined
	\SetKwProg{Fn}{Function}{}{end}
	\KwData{$x_{S}$: a group of tokens of size $S$, \\ 
	\quad \quad $E$: expert network \\
	\quad \quad $T_{S}$: number of iterations for shared-expert, \\
	\quad \quad $T_{D}$: number of iterations for dense-gate, \\
	\quad \quad $T$: number of training iterations.}
% 	\KwResult{$y_{S}$: tokens after transformation of experts}
	\SetInd{0.61em}{0.61em}
	\For{$i \leftarrow 1\ to\ T_{S}$}{
        $y_{S} \leftarrow e(x_{S})$ \;
    }
    // Diversify experts from the shared \;
	\For{$e_{i} \in E$}{
	    $e_{i} \leftarrow diversify(e, i)$ \;
	}        
	\For{$i \leftarrow T_{S}\ to\ T$}{
	$\tau \leftarrow \text{\textit{temperature scheduler}}(i)$ \;
	// Get selected expert ids and weights for each token \;
	$G_{S,\ E}, Id_{S} \leftarrow DTS\_Gate(x_{S},\ \tau, T_{D})$\;
	\For{$s \leftarrow 1\ to\ S$}{
	    $y_{s} \leftarrow 0$
    	\For{$id \in id_{s}$}{
    	    $y_{s} \leftarrow y_{s} + G_{s, \ id} * e_{id}(x_{s})$ \;
    	}
	}    	
	}
\caption{Training MoE in the \evomoe{} Framework}
\label{alg:std_algo}
\end{algorithm}

\subsection{Stage 1: Expert-Diversify}
As the gate network and expert network are both randomly initialized, it requires a vast amount of computation budget for trial and errors, which is inefficient for models' training. Based on the observation in Section~\ref{sec:motivation} that most tokens are processed by the same expert and other experts waste their computation budget, we train one shared-expert instead of $N$ individual experts in the early stage (illustrated as the left of Figure~\ref{fig:overview}). Because all experts within the same MoE layer share weights, the model is equal to its corresponding non-MoE model as a small dense model.

Algorithm~\ref{alg:std_algo} illustrates the MoE training process in our \ourmethods{} framework. First, input tokens are processed by the shared expert $e_{0}$ (line 1-2). Then \ourmethods{} switches the training into standard MoE models' training, by adding a gate network at each MoE layer and diversifying all experts from the shared expert (line 4-5). After this expert-diversify phase, \ourmethods{} steps into the gate-sparsify phase, where it schedules the gate temperature coefficients and then obtains the token-to-expert routing relation from DTS-gate (line 7-9). Tokens will be dispatched to corresponding experts and aggregated together by weighted sum operating (line 10-12).

Multiple $diversify$ techniques can be adopted to spawn multiple diverse experts from one expert, such as noise, NAS, random mask.
\ourmethods{} adopts the random mask, which masks part of the shared expert's weights as 0 (shown as Figure~\ref{fig:shared_to_diverse}). For example, expert-1 is initialized by masking the central value. The proposed expert-diversify stage avoids joint training from scratch and the well-trained diverse experts could be treated as a better initialization to benefit the following model convergence.

\subsection{Stage 2: Gate-Sparsify}
Although sparse gating has demonstrated its superior model efficiency in both training and inference, prior work tends to convergence to a sub-optimal model under the fixed computation budget or the dataset size due to the jointly training of the randomly initialized gate network and expert network.   
In this paper, we propose a new mechanism for training the gate network, named \texttt{Dense-to-Sparse} gate (DTS-Gate, as illustrated in Algorithm~\ref{alg:dts_algo} ), which starts as a dense gate that routes tokens to most experts and then gradually becomes sparser. 
DTS-Gate benefits from the sufficient training of experts in the early stage and then make the experts selection becomes sparser on the basis of specialized experts.
This dense-to-sparse process only occupies a small fraction compared with the total training time, which usually takes days to weeks.

\textbf{Gate with Temperature:}
In order to control the sparsity during training, we adopt the softmax temperature to adjust the weights distribution among experts. Formulated as Equation~\ref{equ:gumbel_softmax}, $W_{g}$ is the parameter of gate, $\zeta$ is the extra noise and sampled from \texttt{Gumbel}$(0, 1)$ distribution~\citep{DBLP:gumbel_softmax}, and $\tau$ is the softmax temperature which controls the distribution. 
When the $\tau$ increases, the distribution of $g'(x_{s})$ becomes more uniform, which evolves more experts into the computation of each token.
As the $\tau$ approaching 0, the distribution becomes one-hot, which is more confident for the gate network.

\begin{equation}
g'(x_{s})  =  \frac {e^{(x_{s} \cdot W_{g}+\zeta)/\tau }}{\sum _ {s'=1}^ {N}e^{(x_{s'} \cdot W_{g}+\zeta)/\tau }}
\label{equ:gumbel_softmax}
\end{equation}

\textbf{Content-based Sparsity:}
Different from existing static Top-K based gate~\cite{DBLP:switch}~\cite{DBLP:gshard}, \ourmethods{} adopts the \textit{content-based sparsity} method to determine the number of activated experts, which keeps the value beyond an threshold $c$. As formulated by Equation~\ref{equ:adaptive_sparsity}, we drop the experts whose weights fall below the threshold $c$ and no extra communication or computation will be wasted. 
It's worth noting that the sum of selected experts' score can not be equal to 1 because we don't normalize them after dropping. It is useful to remain the original score, especially only one expert is selected, which was verified in Switch. To meet the demand of above two designs, we enable each expert with this content-based gate to make them well specialized. transformer~\cite{DBLP:switch}.

\begin{equation}
g(x_{s})_i = \begin{cases}
  g'(x_{s})_i, &\text{if}\quad {g(x_{s})_i} > c \\
  0, &\text{else}
\end{cases}
\label{equ:adaptive_sparsity}
\end{equation}

\textbf{Sparsity Scheduler:} 
With temperature $\tau$ increasing, the distribution tends to be uniform and more experts will be selected. So the sparsity decreases and the training cost of the neural network would increases.
On the opposite, less experts are involved into computation and thus the sparsity increases.
By scheduling the temperature of Equation~\ref{equ:gumbel_softmax}, we can control the sparsity of the MoE layer over different training stages. There is a trade-off between model quality and training cost for the selection of temperature. 
For example, when the distribution of experts is nearly one-hot, it would lead to large variance of gradients between experts and thus make the learning of MoE learning difficult. To optimize this problem, our DTS-Gate starts at a large temperature that routes tokens to most experts and then anneals to a small temperature that gradually sparsifies the MoE layer. 

\textbf{Balance Loss:} Similar to Switch transformer~\cite{DBLP:switch}, we utilize the balance loss $\mathcal{L}_{balance}$ to avoid imbalanced assignments for different experts which would cause the straggler problem and thus lead to low training efficiency.
\begin{equation}
\mathcal{L}_{balance} = \alpha N \sum_{i=1}^N(\frac{\sum_{x_s \in \mathcal{B}}{\mathbb{I}\{g(x_s)_i} > 0\}}{|\mathcal{B}|^2} \sum_{x_s \in \mathcal{B}}{g'(x_s)_i})
\label{equ:balance_loss}
\end{equation}

\noindent As formulated in Equation~\ref{equ:balance_loss}, $\alpha$ is a hyper-parameter and $\mathcal{B}$ represents current batch of tokens. $\sum_{x_s \in \mathcal{B}}{\mathbb{I}\{g(x_s)_i} > 0\}$ represents the number of tokens dispatch to expert $i$ and $\sum_{x_s \in \mathcal{B}}{g'(x_s)_i}$ represents the gate probability allocated for expert $i$. Intuitively, the balance loss will reduce the amount of data for overloaded-experts and move towards to balanced loads  at the batch data level.

\textbf{Training Objective:}
In the first stage, the experts of each MoE layer share same weights and thus the loads can be divided to them equally. The training objective is to optimize the model quality (i.e., $\mathcal{L}_{task}$). In the second stage, both the model quality and training efficiency (i.e., balanced workloads between experts) need to be considered.
\begin{equation}
\mathcal{L} = \begin{cases}
  \mathcal{L}_{task}, &\text{if}\quad $stage = 1$ \\
  \mathcal{L}_{task} + \mathcal{L}_{balance}, &\text{else}
\end{cases}
\label{equ:train_obj}
\end{equation}

\setlength{\textfloatsep}{0.1cm}
\begin{algorithm}[t]
	\SetAlgoLined
	\SetKwProg{Fn}{Function}{}{end}
	\KwData{$x_{S}$: a group of tokens of size $S$, $\tau$: temperature,\\ \quad\quad $T_{D}$: number of iterations of dense-gate.}
	\KwResult{$G_{S,\ E}$: group combine weights, $Id_{S}$: Index of selected experts}
	\SetInd{0.61em}{0.61em}
    \Fn{DTS\_Gate($x_{S},\ \tau$, $T_{D}$):}{
        $g_{S,\ E} \leftarrow gumbel\_softmax(x_{S} \cdot W_{g},\ \tau)$ \;
    	\If{current\_iteration $<$ $T_{D}$}{
    	//select experts for token, N $\geq$ len($id_{s}$) $\geq$1 \;
    	$Id_{S} \leftarrow select\ g_{S,\ E} \textgreater threshold$ \;
    	}
    	\Else{
    	//select Top-1 expert for token, len($id_{s}$) = 1 \;
    	$Id_{S} \leftarrow select\ Top1(g_{S,\ E})$\;
    	}
    	\For{$s \leftarrow 1\ to\ S$}{
        	\For{$id \in ids$}{
                $G_{s,\ id} \leftarrow g_{s,\ id}$ \;      	    
        	}
    	}
    	Return $Id_{S}, \ G_{S,\ E}$ \;
    % 	$G_{S, E} \leftarrow encode(g_{S, E}, idx)$
    }
\caption{Dense-to-Sparse Gate Mechanism}
\label{alg:dts_algo}
\end{algorithm}

\section{Implementation}
\label{sec:impl}
EvoMoE is implemented by adding support for MoE models on FairSeq\footnote{https://github.com/facebookresearch/fairseq}~\cite{FairScale2021}. Meanwhile, EvoMoE proposes several system optimizations, including: 

\textbf{Topology-Aware Hierarchical All-To-All Communication:}
In AllToAll operation, each GPU sends its data to all GPUs (one-for-all) and receives data sent by all GPUs (all-for-one), where each data will be divided equally into n parts.
Current AllToAll operations implemented in NCCL and MPI may suffer from low utilization of network bandwidth because of the small message size.  We propose \texttt{Topology-Aware Hierarchical AllToAll}, which combines hierarchical networks (intra-node and inter-node) and aggregates messages, to optimize the communication between multi-nodes equipped with one NIC. It first gathers the data of eight GPUs inside the same node into one GPU, and then performs a data layout transformation to orignize the placement of tokens. Afterwards, it launches All-To-All communication between nodes, and then performs the data layout transformation and scatters tokens to its corresponding expert. In this way, the size of data transferred between nodes is ${\#\text{GPU}}^2$ times larger than before. 
Meanwhile, this two-level decoupled AllToAll also fully utilizes the intra-node (NvLink or PCIe) and inter-node bandwidth (Infiniband or Ethernet).

\textbf{MoE-Aware Recomputation:} Recomputation is a mainstream techniques to reduce the memory footprint of training models, which evicts feature map tensors in the forward pass for memory saving and then regenerates them for calculating gradients by executing corresponding computation. Existing systems, e.g., DeepSpeed~\cite{DBLP:conf/kdd/deepspeed} and FairSeq~\cite{DBLP:conf/naacl/fairseq}, 
adopt recomputation as the recommended configuration for training large models, which only saves the input tensor of each Transformer layer and re-executes the whole Transformer in the backward. As the MoE layer involves two All-To-All communication in the forward, re-executing them may lead to large time cost. To keep the memory efficiency while improve training efficiency, we propose the MoE-aware recomputation, which save three tensors of each Transformer layer, including the input tensor of Transformer layer, and the input tensor of two All-To-All operations.

\section{Experiments}
\label{sec:experiment}
\begin{table*}[t]
    \caption{Evaluating the pre-trained models on the \textbf{GLUE Benchmark}. We scale the Standard Transformer(TRM) into the MoE model by replacing every the other FFN with a MoE layer.}
        \begin{center}
            \begin{tabular}{l|ccc|cccccccc|c}
            \toprule 
            \textbf{Models} & \textbf{\#Shared Params.} & \textbf{\#Expert Params.} &\textbf{FLOPs}& \textbf{MNLI} & \textbf{QNLI} & \textbf{QQP} & \textbf{RTE} & \textbf{SST-2} & \textbf{MRPC} & \textbf{CoLA} & \textbf{STS} & \textbf{Avg} \\
            \midrule
            Standard TRM& 355M & -& 207B & 88.2& 93.2& 92.1 & 85.1 & 95.8 & 88.6 & 84.5 & 90.5 & 89.750\\
            Larger TRM& 370M &  -&  220B& 88.4& 93.5& 92.2 & 85.3 & 95.8 & 88.9 & 84.7 & 90.6 & 89.925\\
            \midrule
            Switch TRM& 259M & 1536M & 220B&89.2& 93.7 & 92.2 & 86.4& 95.8 & 89.1& 85.3 & 90.8 & 90.313\\
            BASE Layer & 259M &1536M & 220B&89.5& \textbf{93.9}& \textbf{92.4}& 87.3 & 96.0 & 89.4 & 85.5 & 91.2 & 90.650\\
            Hash Layer&259M &1536M & 220B&89.4& \textbf{93.9}& 92.2 & 87.1 & 95.9 & 89.2 & 85.5 & 90.9 & 90.513\\
            StableMoE &259M &1536M & 220B&89.3& 93.8 & 92.1 & 86.7& 95.8& 89.2 & 85.4 & 91.0 & 90.413\\
            EvoMoE &259M &1536M & 220B& \textbf{89.9} & \textbf{93.9} & 92.3 & \textbf{88.1} & \textbf{96.1} & \textbf{89.6} & \textbf{85.6} & \textbf{91.5} & \textbf{90.875}\\
            \bottomrule
            \end{tabular}
        \end{center}
\label{tab:roberta_glue}
\end{table*}

\subsection{Experimental Setup}

\subsubsection{Machine Environment}
We conduct experiments on DGX-A100, where each server is equipped 2 AMD CPUs and 8 NVIDIA Ampere A100 (40GB) GPUs, with Ubuntu 20.04, CUDA 11.3, CuDNN 8.2.0 and NCCL 2.12.7. GPUs inside a server are connected via NVLink 3.0
with and servers are connected with 8 InfiniBand NICs via 8*200 Gbps bandwidth totally. The RDMA is used by default and the PyTorch version is 1.11. \\ \\

\subsubsection{Baselines}
To verify the effectiveness of our method, we compare it with the several representative baselines, including Switch-Transformer~\cite{DBLP:switch}, BASELayer~\cite{DBLP:conf/icml/baselayer}, HashLayer~\cite{roller2021hash}, DSelectK~\cite{DBLP:conf/nips/dselectk}
and StableMoE~\cite{DBLP:conf/acl/stablemoe}.
Switch-Transformer~\cite{DBLP:switch} proposed to adopt Top-1 gate for the training of large-scale models.
BASELayer~\cite{lewis2021base} formulates the token-expert routing as a linear assignment problem and guarantees balanced compute loads by employing numerous algorithms. 
HashLayer~\cite{roller2021hash} replaces the gating network with a hash-based routing strategy (e.g., random hash, clustered hash). DSelectK~\cite{DBLP:conf/nips/dselectk} involves sparse gates (Top-K) in the multi-gate MoE (i.e., MMoE) architecture for better parameter sharing among different tasks and trains gates from dense to sparse for smoothness. StableMoE~\cite{DBLP:conf/acl/stablemoe} also proposed two training stages, which learn the gate as well as distill it into a lightweight one in the first stage, and freezes the parameter of gate for stable routing in the second stage.
Our \evomoe mainly contains two phases: an \texttt{expert-diversify}
%spawning 
phase to spawn multiple diverse experts from one single well-trained base expert,
and a \texttt{gate-sparsify} phase that gradually and adaptively learns a increasingly sparse gate from a dense gate.

\subsubsection{Benchmark and Datasets}
We evaluate \evomoe{} on three popular tasks, including the machine translation (MT) task for domain-specific models, the Masked Language Modeling (MLM) task and the language modeling (LM) task for pre-trained models.

We adopt standard Transformer architecture~\cite{vaswani2017attention} (Encoder-Decoder) 
for the MT task and train models on four popular translation datasets, WMT17 (English to German/German to English)\footnote{https://www.statmt.org/wmt17/}, and WMT14 (English to French/French to English)\footnote{https://www.statmt.org/wmt14/}. BLEU scores of the test sets are reported for comparison.

We adopt RoBERTa architecture architecture~\cite{liu2019roberta} (Encoder-only) 
for the MLM task and train models on the combination of datasets, including Wikipedia\footnote{https://dumps.wikimedia.org/enwiki/latest/enwiki-latest-abstract.xml.gz}, BooksCorpus\footnote{https://battle.shawwn.com/sdb/books1/books1.tar.gz}, OpenWebText\footnote{https://zenodo.org/record/3834942/files/openwebtext.tar.xz} and CC-100\footnote{https://data.statmt.org/cc-100/}. Moreover, these datasets are tokenized by byte-pair encoding with a vocabulary size of 50257. Models are validated on the famous General Language Understanding Evaluation(GLUE) benchmark~\cite{DBLP:conf/emnlp/glue} for comparison.

We adopt GPT architecture architecture~\cite{gpt2} (Decoder-only) 
for the LM task and train models on OpenWebText as ~\citet{gpt2}. We report train/valid/test perplexity (PPL) for comparison.

We also report the inference FLOPs of each model, which represents the speed of deploying this model at industry. All the training data are downloaded and pre-processed by following the example scripts from Fairseq\footnote{https://github.com/facebookresearch/fairseq/tree/main/examples}.

\subsubsection{Hyper-Parameter Detail}
We sparsely scale these models by replacing every other the \textit{feed-forward} layer (FFN) with MoE-FFN Layer, which contains a series of FFN experts.
All models use the GeLU activation functions~\cite{gelu_activation}, polynomial learning rate scheduler and Adam optimizer~\cite{DBLP:journals/corr/adam_optim}, where $\beta_1 = 0.9$ and $\beta_2 = 0.98$ . 
We set clip norm as 0.0, weight decay as 0.1 and dropout rate as 0.1.
We use CrossEntropy as the criterion and utilize the label smoothed technique with coefficient of 0.1 for the MT task.  
The coefficient of balance loss is set as 0.1 in Switch-Transformer~\cite{DBLP:switch}, StableMoE~\cite{DBLP:conf/acl/stablemoe} and our EvoMoE.
We set the threshold $c$ of our dense-to-sparse gate as 0.001 over training steps, which determines how large the expert's weight is important and is a trade-off between training cost and model quality from our point of view. 

\begin{table*}[t]
    \caption{Perplexity results of language modeling task.}
        \begin{center}
            \begin{tabular}{l|cccc|c}
            \toprule 
            \textbf{Models}  & \textbf{\#Shared Params.} &  \textbf{\#Expert}& \textbf{\#Expert Params.} & \textbf{FLOPs} & 
            \textbf{Perplexity($\downarrow$)}  \\
            \midrule
            Standard TRM&  345M& -  & -  & 207B& 15.14\\
            Larger TRM (wider)& 360M & - & - & 220B& 14.92\\
            \midrule
            Switch TRM & 249M& 192& 1536M& 220B& 13.12\\
            BASE Layer & 249M& 192& 1536M& 220B& 12.45\\
            Hash Layer& 249M& 192& 1536M& 220B& 12.87\\
            StableMoE & 249M& 192& 1536M&220B & 12.91\\
            EvoMoE & 249M& 192& 1536M& 220B& \textbf{12.24}\\
            \bottomrule
            \end{tabular}
        \end{center}
\label{tab:gpt_lm}
\end{table*}

\begin{table}[t]
    \caption{BLEU score on each machine translation datasets}
        \begin{center}
            \begin{tabular}{c|cccc}
            \toprule 
            \textbf{Models}
            %& {$En-Vi$}  &{$Vi-En$}  
            & \textbf{En-De} & \textbf{De-En} & \textbf{En-Fr} & \textbf{Fr-En}\\ 
            \midrule
            {TRM-Base}  
            & {28.1} &{34.8}
            & {39.2} &{38.1}\\
            {Switch-TRM}
            & {28.4} &{34.6}  
            & {39.1} &{38.2} \\
            \midrule
            {\evomoe{}} 
            & {\textbf{29.6}} &{\textbf{36.7}}
            & {\textbf{40.3}} &{39.2} \\
            {\quad\quad\quad\quad w/o \ stage\ 1}  
            & {\textbf{29.6}} &{36.5}  
            & {40.2} &{\textbf{39.3}} \\
            {\quad\quad\quad\quad w/o \ stage\ 2}  
            & {28.7} &{35.2}  
            & {39.4} &{38.3} \\
            \bottomrule
            \end{tabular}
        \end{center}
\label{tab:trans_tasks}
\end{table}

\subsection{GLUE Results}
\textbf{Model Architecture:} We pretrain the representative RoBERTa model for the masked language modeling task, where we set standard Transformer(TRM) with 24 encoder layers, hidden dimension as 1024 and number of attention heads as 16. We replace every other FFN layer in standard Transformer with the MoE layer (16 experts per layers) to construct the MoE models.  The standard Transformer is a dense model and contains 355M parameters totally, whose inference FLOPs is 207B. Meanwhile, the sparse MoE model contains 1759M parameters totally, including 259M parameters for shared backbone and 1536M parameters for the expert network. In our setting that only 1 expert is active at a time, each input token will activate 335M parameters of the sparse MoE models, which is the same as standard Transformer model except for the gate network. To exactly match the inference speed (FLOPs) of MoE models, we slightly increase the FFN hidden size of standard TRM to construct the larger TRM.

\textbf{Model Performance:}
We pretrained each model for 100k steps totally, 5k of which was the warm-up phase. For our \ourmethods{}, we scheduled the first 5k steps as the expert-diversify stage and the following 5k steps for annealing temperature from 2.0 to 0.3. After the pre-training stage, we finetune the pre-trained models on each GLUE task and summarized the results in Table~\ref{tab:roberta_glue}. As for RTE, we finetune it starting from the MNLI model rather than the pretrained model as ~\citet{liu2019roberta}. 

Compared with other baselines, \ourmethods{} achieves state-of-the-art results on 7 out of 8 tasks and the best averaged score. The MoE model is constructed by adding the gate network and replacing the original FFN layer of standard Transformer, which increases its model size and thus enlarges its capacity. Thus all MoE models outperform their backbone model (standard TRM), e.g., 89.750 for standard TRM and 90.313 (+ 0.563) for Switch TRM with respect to the avg score. Larger TRM slightly outperforms standard TRM because of its large model size. As verified by ~\citet{DBLP:scaling_laws}, larger models tend to be more sample-efficient, which represents better model quality with fixed training data/steps.

Compared with other MoE methods, \ourmethods{} benefits from the sufficient training of experts in the early stage and then make the experts selection becomes sparser on the basis of specialized experts. 
Specifically, \ourmethods{} outperforms other MoE methods on GLUE benchmark up to 0.562 and 0.403 on average. 
Switch TRM~\cite{DBLP:switch} jointly trains the randomly initialized experts and gates, which aims to learn better parameter as well as balanced routing. It is hard to optimize them simultaneously and thus performs bad among MoE models. To alleviate this problem, StableMoE~\cite{DBLP:conf/acl/stablemoe} freezes the parameter of gate network after the early training stage and improves over Switch-TRM. Hash Layer~\cite{roller2021hash} utilizes the fixed hash strategy to route tokens, which is based on the input embedding. Because both the hash strategy and input embedding is fixed, Hash Layers only need to learn the parameter of experts. However, it may lead to sub-optimal because the hash strategy is selected based on human knowledge and may be inappropriate. BASE Layer~\cite{DBLP:conf/icml/baselayer} enforces a balanced token-to-expert assignment through a linear assignment problem, which simplify the training in another way. All these work find the problem of jointly training and targeting at alleviate it.

\subsection{Language Modeling Results}
\textbf{Model Architecture:} We pretrain the representative GPT model for the language modeling task, where we set standard Transformer(TRM) with 24 decoder layers, hidden dimension as 1024 and number of attention heads as 16. 
Every other FFN layer is replaced by in standard Transformer with the MoE layer (16 experts per layers) to construct the MoE models. There totally exists 12 MoE layers and thus 192 experts (i.e., $12 \times 16$).  
Meanwhile, larger TRM is scaled by increasing its FFN hidden size.

\textbf{Model Performance:}
We pretrained each model on the OpenWebText dataset for 200k steps totally, 10k of which was the warm-up phase. For our \ourmethods{}, we scheduled the first 10k steps as the expert-diversify stage and the following 5k steps for annealing temperature from 2.0 to 0.3. We report the perplexity on the test set. Results are summarized in Table~\ref{tab:gpt_lm}.

Compared with other baselines, our \ourmethods{} achieves the best result among all baselines. Specifically, the perplexity of \ourmethods{} is 12.24, which achieves a 2.90 improvement compared with 15.14 of standard TRM. Meanwhile, all MoE models outperform their backbone model (standard TRM) because of their large model capacity. Larger TRM slightly outperforms standard TRM because of its large model size, which demonstrates the sample-efficient of large models.

Compared with other MoE methods, \ourmethods{} benefits from the sufficient training of experts in the early stage and then make the experts selection becomes sparser on the basis of specialized experts. Specifically, \ourmethods{} outperforms other MoE methods up to 0.88 ppl and 0.545 ppl on average. The analysis between different methods is same as that in GLUE results.

\begin{table}[t]
    \caption{BLEU performance of MoE models with different expert number. }
        \begin{center}
            \begin{tabular}{c|ccc}
            \toprule 
            {} & \multicolumn{3}{c}{\textbf{Number\ of\ Experts }} \\
            &{\textbf{4}} & {\textbf{8}} &{\textbf{16}}\\ 
            \midrule
            {Switch}
            &{28.4} 
            &{28.6} &{28.7}\\
            {\ourmethods{}} 
            &\textbf{29.6} &\textbf{29.9}
            &{30.0} \\
            {\quad\quad\quad\quad w/o \ stage\ 1}  
            &\textbf{29.6} &\textbf{29.9}  
            &\textbf{30.1}  \\
            {\quad\quad\quad\quad w/o \ stage\ 2}  
            &{28.7} &{28.9}  
            &{28.9} \\
            \bottomrule
            \end{tabular}
        \end{center}
\label{tab:expert_number}
\end{table}

\begin{table}[t]
    \caption{Efficiency of MoE models with different expert number, and the results are normalized over Switch.}
        \begin{center}
            \begin{tabular}{c|ccc}
            \toprule 
            {} & \multicolumn{3}{c}{\textbf{Number\ of\ Experts }} \\
            &{\textbf{4}} & {\textbf{8}} &{\textbf{16}}\\ 
            \midrule
            {Switch}
            &{1} 
            &{1} &{1}\\
            {\ourmethods{}} 
            &\textbf{0.82}
            &\textbf{0.78} &\textbf{0.75} \\
            {\quad\quad\quad\quad w/o \ stage\ 1}  
            &{0.86}  
            &{0.82} &{0.81} \\
            {\quad\quad\quad\quad w/o \ stage\ 2}  
            &{0.95}  
            &{0.93} &{0.92} \\
            \bottomrule
            \end{tabular}
        \end{center}
\label{tab:efficiency}
\end{table}

\begin{figure*}[t]
    \begin{center}
    \subfigure[Validation PPL over steps]{
            \includegraphics[width=0.35\textwidth]{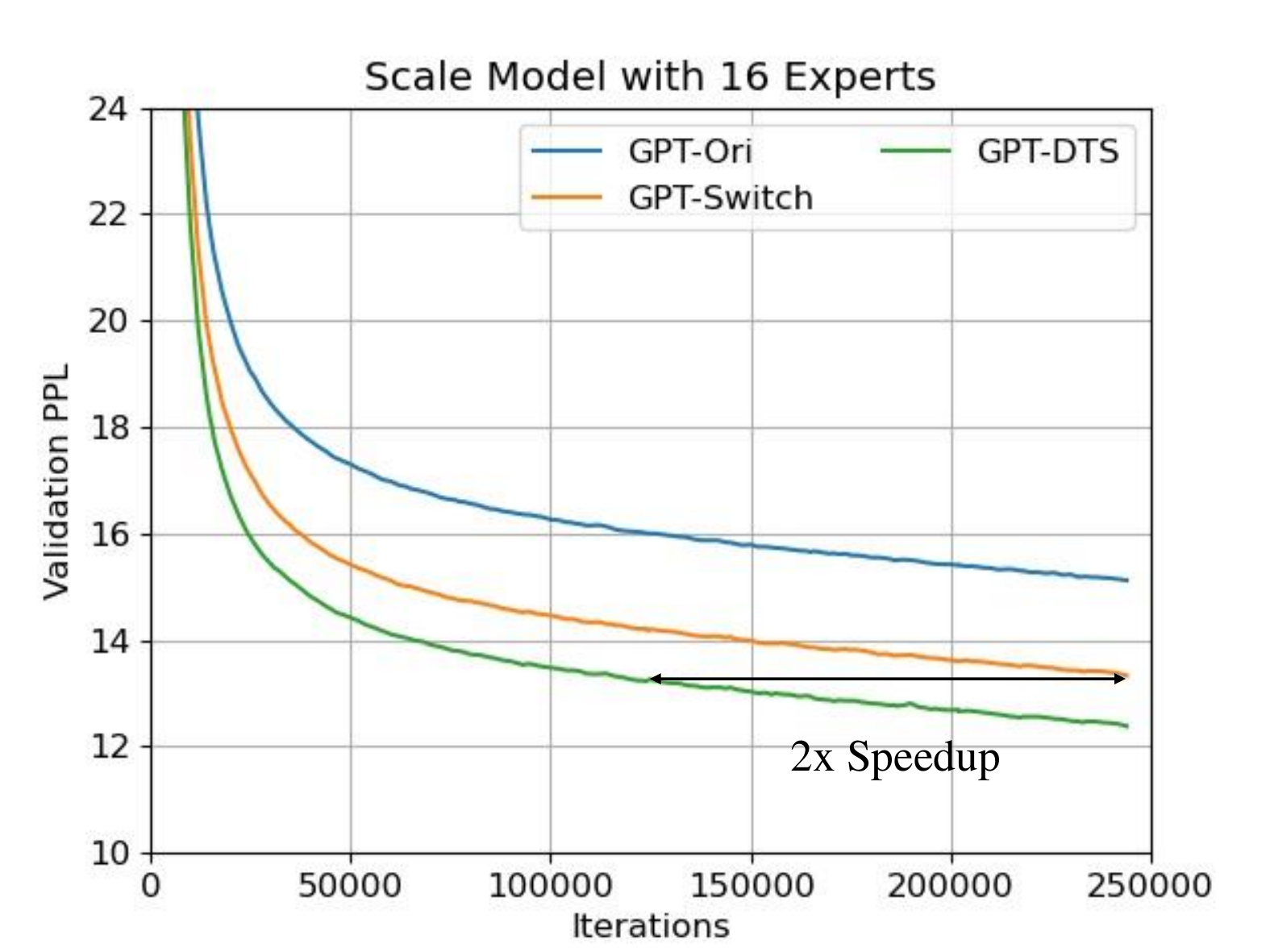}
            \label{fig:iters_ppl}        }    
    \subfigure[Validation PPL over FLOPs]{
            \includegraphics[width=0.35\textwidth]{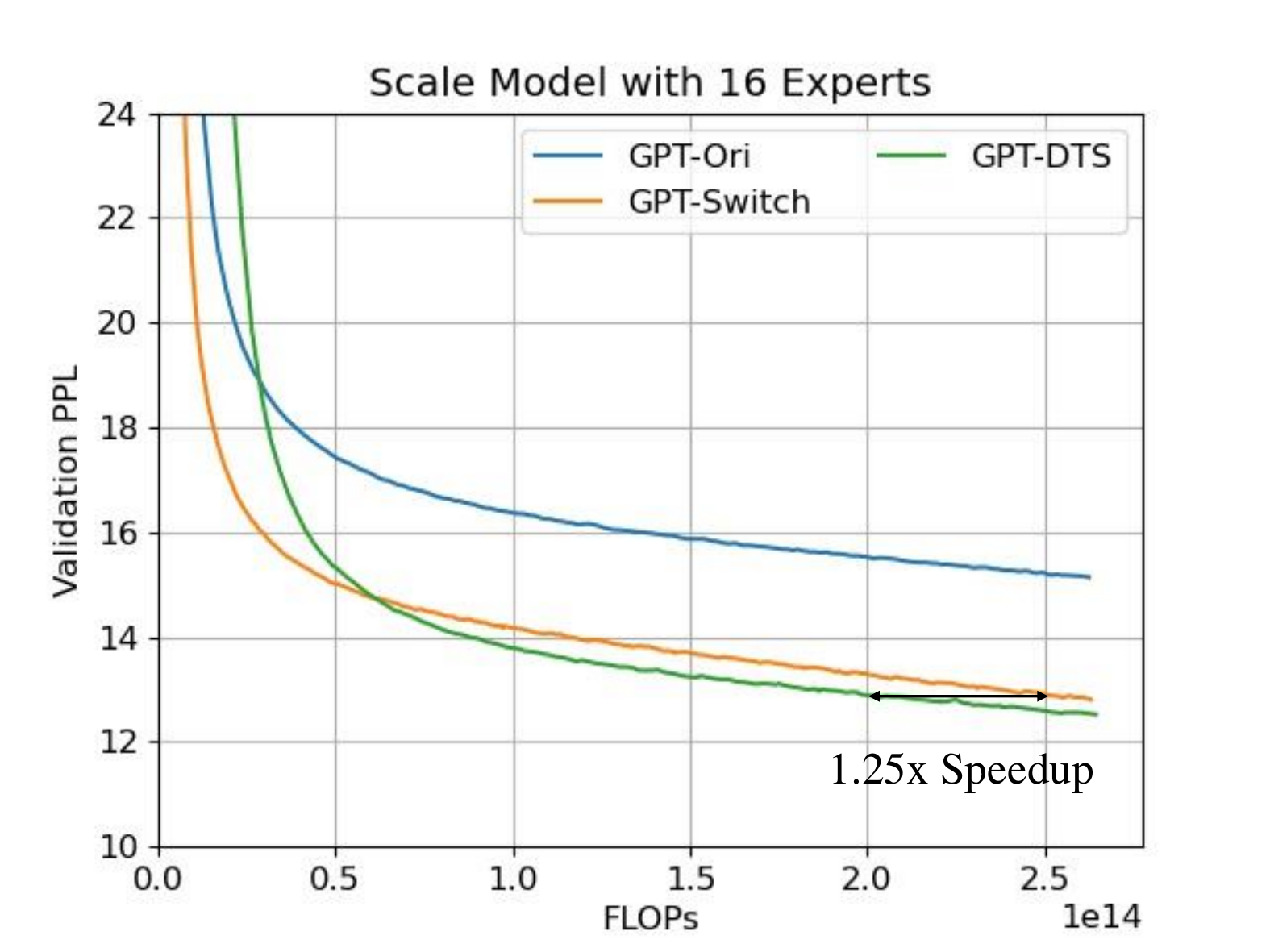}
            \label{fig:flops_ppl}
    }
    \caption{End-to-end performance comparison between GPT-ori, GPT-Switch and GPT-DTS. Figure~\ref{fig:iters_ppl} and Figure~\ref{fig:flops_ppl} represent the curve of PPL over iterations and FLOPs, where GPT-DTS can obtain $2.0$x speed-up to reach the same validation perplexity, as well as higher FLOPs-efficiency of a $1.42$x speed-up.
}
    \label{fig:end2end}
    \end{center}
\end{figure*}

\subsection{Machine Translation Results}

\textbf{Model Architecture:} We pretrain the representative T5 model for the machine translation task, where we set standard Transformer(TRM) with 12 encoder-decoder layers, hidden dimension as 768 and number of attention heads as 12. 
Every other FFN layer is replaced by in standard Transformer with the MoE layer (4 experts per layers) to construct the MoE models. 

\textbf{Model Performance:}
We compare \evomoe{} with Transformer and Switch-Transformer on four language-pair datasets, including English to German, German to English, English to French and French to English. Results are shown in Table~\ref{tab:trans_tasks},~\ref{tab:expert_number},~\ref{tab:efficiency}.
We remark that these models all have the same inference speed even if MoE models enlarge the parameter size.
We show the BLEU score on the test set of each dataset in Table~\ref{tab:trans_tasks}.
\ourmethods{} outperforms other methods by 1 BLEU score on average.
Although Switch-Transformer scale the model size, it still achieves a similar performance with Transformer-Base, which is parameter-efficient.
Table~\ref{tab:expert_number} shows the BLEU performance of different expert number on the English-German datasets. \ourmethods{} can still outperform the Switch-Transformer about 1.3 BLEU with the increasing number of experts. Because of the datasets' quality, the effect of increasing expert number is limited.

\textbf{Model-Efficency:} Table~\ref{tab:efficiency} show the model efficiency of \ourmethods{} and Switch Gate on the English-German datasets. \ourmethods{} is efficient at the speed of converge. For example, \ourmethods{} need only $75\%$ compute budget of Switch-Transformer to reach the same PPL. It is worth noting that the speedup over Switch-Transformer improves as the expert number increases.

\textbf{Ablation Study:}
We present a ablation study on \ourmethods{} to show the influence of two stages by removing the expert-diversify phase and the gate-sparsify phase respectively. Results are summarized in Table~\ref{tab:trans_tasks}~\ref{tab:expert_number}~\ref{tab:efficiency}.
As for the model performance metric, it will lead to performance degradation when \ourmethods{} removes the gate-sparsify stage, such as 38.3/39.2 in Fr-En of Table~\ref{tab:trans_tasks}. 
Meanwhile, it is worth noting that influence is little as for w/ and w/o the expert-diversify stage, which encourages us to involve this stage for saving computation budget.
As for the FLOPs-efficiency metric, the gate-sparsify phase can improve the FLOPs-efficiency by $17\%$. By introducing the expert-diversify stage, \ourmethods{} can obtain an extra $4\%$ improvement.

In summary, the gate-sparsify phase can both improve the model performance and FLOPs-efficiency significantly and the expert-diversify phase can introduce extra FLOPs-efficiency without performance degradation. In the following sections, we will detail analyze the gate-sparsify phase and evaluate it at large scale.

\subsection{Breakdown on the Gate-Sparsify Phase}
% Detailed analysis about Dense-to-Sparse Gate.

\textbf{Model Architecture:} We pretrain the representative GPT model for the language modeling task, where we set GPT-ori with 24 decoder layers, hidden dimension as 1024 and number of attention heads as 16. 
Every other FFN layer is replaced by in standard Transformer with the MoE layer (16 experts per layers) to construct the GPT-MoE model. 
GPT-Switch represents training MoE models with Switch gate, which keeps Top-1 selection.
GPT-DTS represents training MoE models with the \texttt{dense-to-sparse} gate, which starts as a dense gate that routes tokens to most experts and then gradually becomes sparser. 

We compare the required FLOPs to train models to show the FLOPs-efficiency of different methods.
The FLOPs-efficiency is defined as the best model performance (PPL) can be achieved given the fixed number of floating-point operations (computation budget).
Because the actual training time could be affected by the system engineering efforts on the implementation details, which are not our focus in this approach. Instead, in our experiments, we prefer to choose the computation complexity for fair comparisons. 

\begin{figure*}[t]
    \centering
    \subfigure[More Experts]{
            \includegraphics[width=0.33\textwidth]{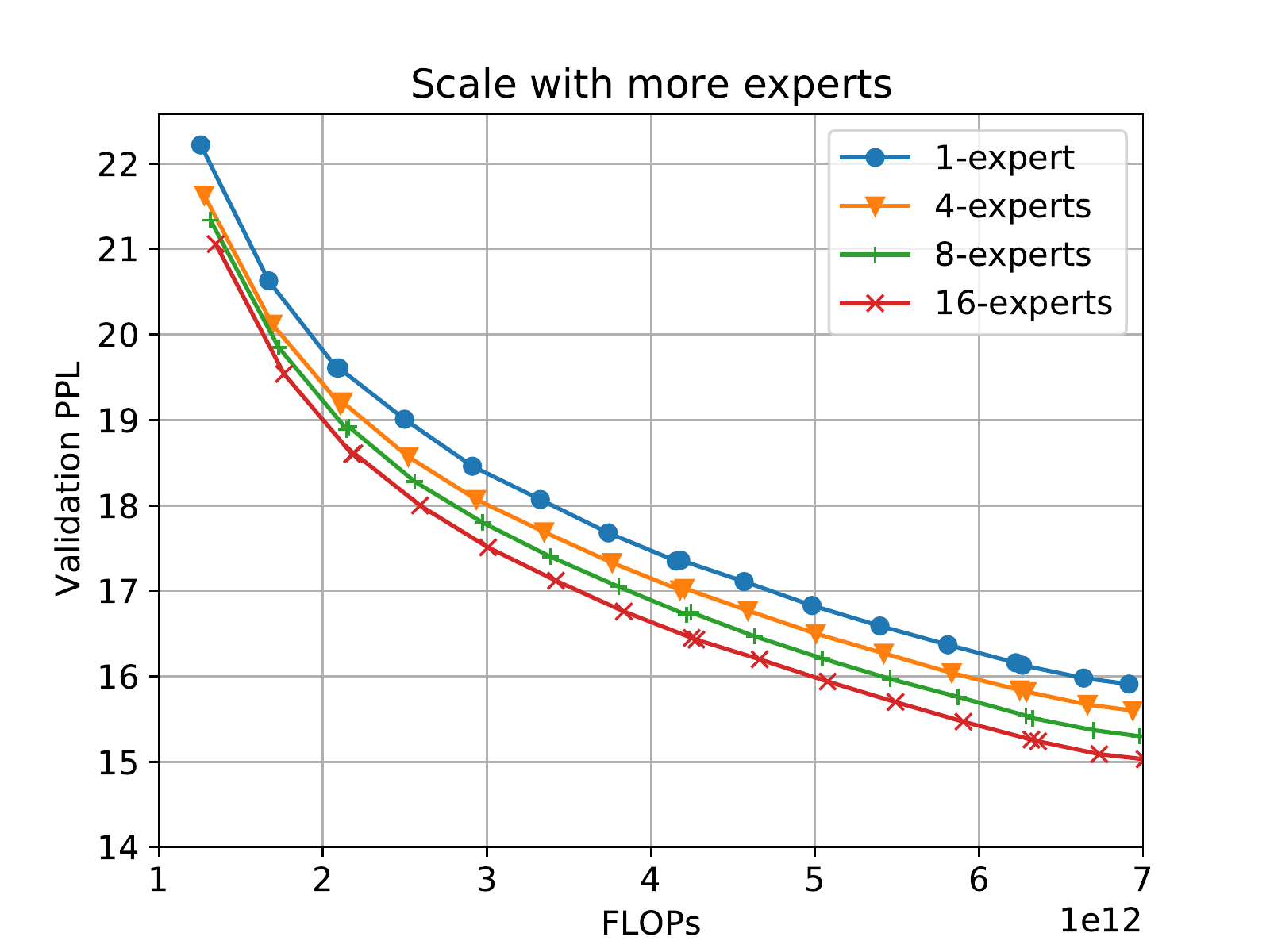}
            \label{fig:expert_scalability}
    }
    \subfigure[More MoE Layers]{
            \includegraphics[width=0.33\textwidth]{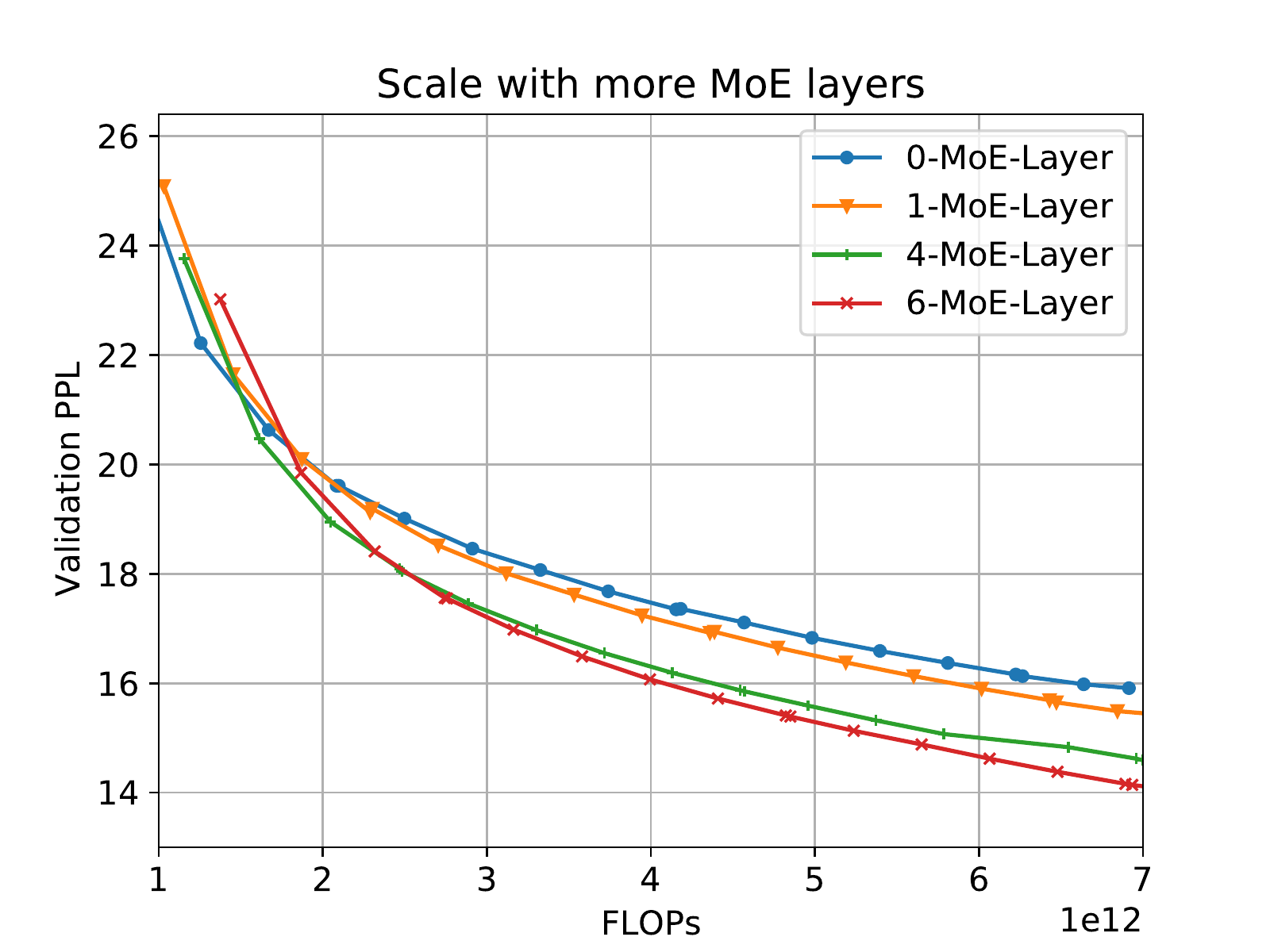}
            \label{fig:moe_layer_scalability}
        }
    \caption{Scalability for DTS gate. It shows that more experts or more MoE-layers (larger models with constant FLOPs), will lead to better FLOPs-efficiency.
    }
    \label{fig:scalability}
\end{figure*}

\textbf{Model Performance:}
We pretrained each model on the OpenWebText dataset for 300k steps totally, 10k of which was the warm-up phase. For our \ourmethods{}, we scheduled the first 10k steps for annealing temperature from 2.0 to 0.3. We report the perplexity on the validation set. Results are shown in Figure~\ref{fig:end2end}.
To improve the computation efficiency, only part parameters are used for each token in sparse models with the cost of model performance. 
DTS-Gate aims to shift the model training from dense to sparse, and keep the inference cost same as sparse models.
Experiments show that compared with the state-of-the-art Switch-Transformer in GPT-MoE model with OpenWebText dataset, GPT-DTS can obtain
2.0x speed-up to reach the same validation perplexity (Figure~\ref{fig:iters_ppl}), as well as higher FLOPs-efficiency of a 1.42x speed-up (Figure~\ref{fig:flops_ppl}).
Experiments also verify the ability of \texttt{dense-to-sparse} gate for scaling models with more experts or more MoE layers.

Comparison with Sparse Models.
MoE-Switch pre-defines its static Top-1 gating network and jointly training the gate and experts networks. Different from GPT-Switch, GPT-DTS utilizes temperature to adjust the distribution of the token-to-experts (one-hot or uniform) and threshold to remove computation of experts with low weights. \ourmethods{} performs better than GPT-Switch in sample-efficiency because of more experts involved in training and updates at the beginning, shown as Figure~\ref{fig:iters_ppl}. As for FLOPs-efficiency, DTS-Gate first involves more experts into warm-up training, which is poor FLOPs-efficency. But with the training going on, GPT-DTS can obtain greater than $25\%$ improvements in FLOPs-efficiency compared with the state-of-the-art Switch-Transformer in GPT-MoE model with OpenWebText dataset.

\subsection{Scalability}

In this subsection, we investigate different experiment settings to validate the scalability of our DTS-Gate. 
% and FLOPs-efficiency

\textbf{Model Architecture:} We choose the GPT-small as the backbone model for the language modeling task, where we set the model with 12 decoder layers, hidden dimension as 768 and number of attention heads as 12. 

\textbf{Increase the Expert Number:} Based on GPT-Small model with 117M parameters, we replace the 7-th FFN layer by one MoE layer and vary its experts number within $\{1, 4, 8, 16\}$. 
As shown by Figure~\ref{fig:expert_scalability}, with increasing expert numbers, \ourmethods{} keeps consistent improvements (i.e., lower PPL) during training.

\textbf{Increase the MoE layer number:} Similarly, we also vary the number of MoE layers to validate the performance of DTS gate. 
We insert $k$ MoE layers in GPT-Small, where $k \in \{0, 1, 4, 6\}$ and each MoE layer contains 8 experts. 
Figure~\ref{fig:moe_layer_scalability} shows that by increasing MoE layers, \ourmethods{} can achieve better model performance with same FLOPs.

\subsection{Effect of Sparsity Scheduler}

It is worth noting that several hyper-parameters are introduced in Dense-To-Sparse gate, such as max/min temperature and decay iterations.
In this section, we analyze the effect of different hyper-parameter setting by conducting  experiments of various settings. The training model is GPT-MoE, 24-layer decoder with 12 MoE-layer (16 experts per layer) and the dataset is OpenWebText.

\begin{figure}[t]
    \centering
    \includegraphics[width=0.35\textwidth]{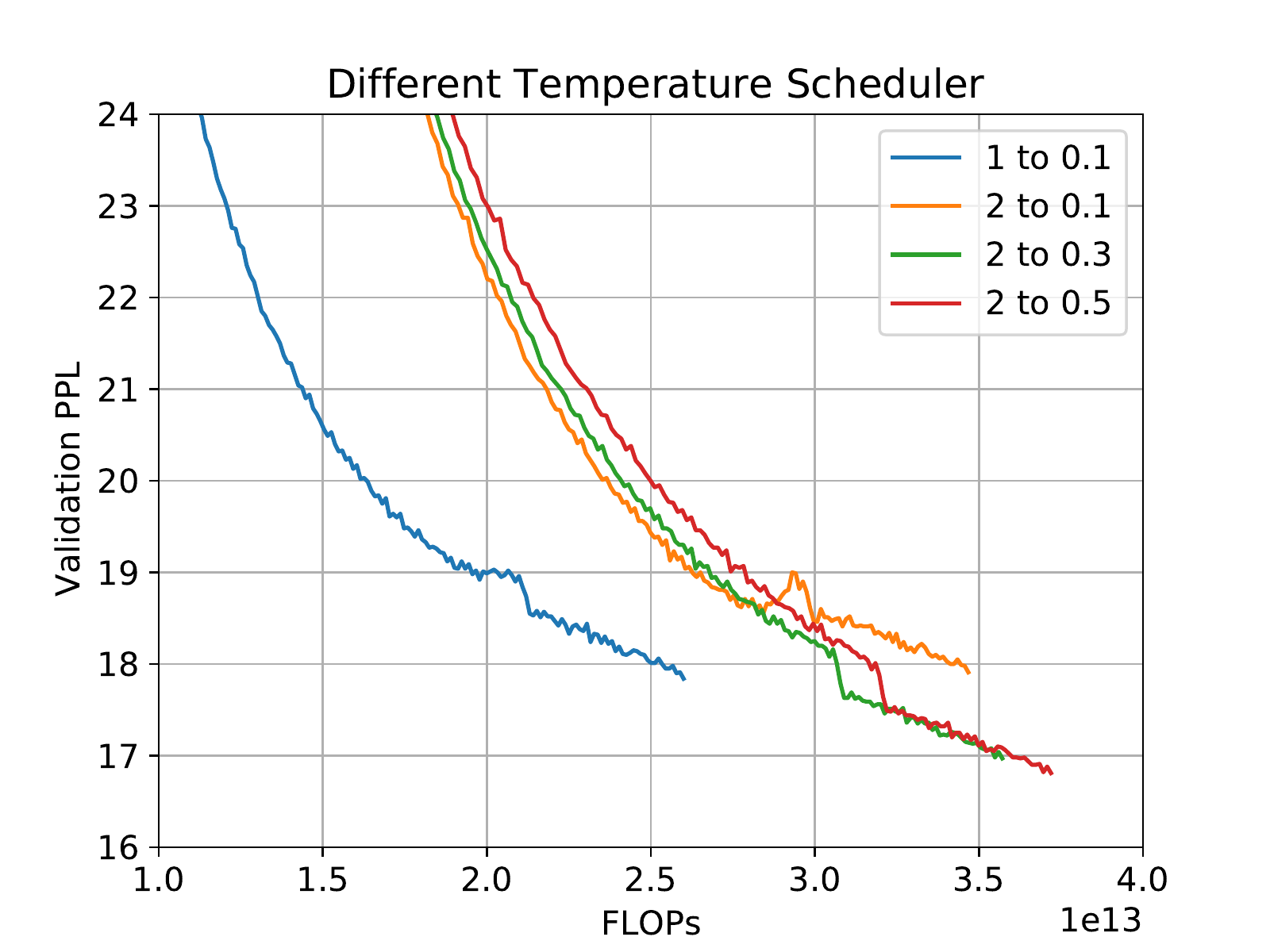}
    \caption{Effect of different temperature scheduler.}
    \label{fig:scheduler}
\end{figure}

We decay temperature $\tau$ from $max\_value$ to $min\_value$ in the first 15000 iterations and switch to Top1 then. Experiments with different $max\_value$ to $min\_value$ are evaluated, and the results are shown in Figure~\ref{fig:scheduler}.

\textbf{Max/Min Temperature} Under small temperatures, the weight distribution of experts is close to one-hot, which leads to the one-token-one-expert distribution and low training cost, 
but the variance of gradients is large.
In contrast, large temperatures result in nearly uniform distribution gate weights, 
which evolves more experts into training but the variance of gradients is small. As shown in Figure~\ref{fig:scheduler}, we find the these two hyper-parameters have low influence on the model quality under the same training budget, except for the extrame value, e.g., 1.0 for $max\_value$ and 0.1 for $min\_value$.

\begin{table}[t]
    \caption{The most frequent tokens assigned to each expert in the validation set, which shows that some experts assignment decisions are made based on local contexts. For many other experts, the assignment decision depends on longer context, and is harder to visualize.}
        \begin{center}
            \begin{tabular}{c|ccccc|c}
            \toprule 
            \textbf{Expert}  & \multicolumn{5}{c|}{\textbf{Top5 Proceeding Tokens}} & \textbf{Descriptions} \\  
            \midrule
            1 & is & was & be & are & have & auxiliary verbs \\
            3 & .& \textbackslash n & /& (&;  & punctuations \\
            4 & in& , & of& and& from & prepositions \\
            6 & and& I & it& that& they & possessive cases\\
            12 & out& up & go& back& down & directional prepositions \\
            \bottomrule
            \end{tabular}
        \end{center}
\label{tab:visual}
\end{table}

\subsection{Visualization of Expert Specialization}

We visualize the routing strategy of the pre-trained GPT model by \ourmethods{} in Table~\ref{tab:visual} through its corresponding input embedding, where each MoE layer contains 16 experts. For each expert, we present the Top5 proceeding tokens assigned and give descriptions for explanations from our points of view. For example, Expert 1 captures the auxiliary verbs and Expert 6 captures possessive cases. These experts can capture local contexts of each embedding well. For other experts, it is diff cult to visualize because of the long contexts' influence.

\section{Related Work}
\label{sec:related_work}
\paragraph{Static Sparse Neural Networks}
Exploiting the sparsity in deep neural networks can reduce both storage and computation requirements for model training and inference.
One of the most widely used sparsification methods is weight pruning \citep{lecun1990optimal,han2015deep}.
Previous studies proposed to prune away redundant or less useful weights based on various pruning criteria (e.g., the importance of individual weights \citep{han2015deep} or groups of weights \citep{wen2016learning,luo2017thinet,he2017channel}) and then fine-tune remaining weights to regain the lost accuracy.
After pruning and fine-tuning, parts of weights are permanently removed, inducing a static sparsity pattern in DNNs.
The sparsity pattern/structure is a trade-off between model effectiveness and hardware efficiency \citep{mao2017exploring}. %affects all or is a trade-off between 
Early work attempts to increase the sparsity ratio or model accuracy by employing unstructured sparsification methods, while recent work focuses more on structured sparsity for practical speedup on hardware.
Interestingly, \cite{frankle2018lottery} points out training a sparse network from scratch is superior or comparable to pruning-based methods.
Our \evomoe{} adopts a dense-to-sparse gate, which is analogous to pruning-based methods that train all experts first and then learning the sparse gate routing.

\paragraph{Conditional Computation with \moe{}}
Different from previous static sparse neural networks that permanently remove partial weights, conditional computation~\cite{DBLP:journals/corr/BengioLC13} only activates the relevant parameter of the model on a per-sample basis, which can be regarded as a dynamic sparsity structure that remains all model weights but brings sparsity into the computation. %or operation
The mixture-of-expert (\moe{}) architecture~\cite{DBLP:lstm_moe}, as a specific form of conditional computation, contains a series of experts and a trainable gating network which routes each input sample to its corresponding experts.
Conditional computation is capable of reducing inference cost (without reducing model capacity) or increasing model capacity (without increasing inference cost) from a model acceleration or scaling perspective.
On the other hand, the sparsely activated parts (i.e., \moe{} in models) can be regarded as structured sparse blocks, which does not introduce additional computational overhead.
However, conditional computation models are often difficult to train, since they require learning discrete routing decisions from individual examples to experts and the gating network tends to converge to a state that only selects the same few experts~\cite{eigen2013learning}.
LSTM-MoE~\cite{DBLP:lstm_moe}, GShard~\cite{DBLP:gshard} and Switch-Transformer~\cite{DBLP:switch} utilize auxiliary load-balancing losses to mitigate this self-reinforcement phenomenon and thus improve training efficiency.
In such \moe{} models, the gating network and experts, as two critical components,
are jointly trained which may interfere with each other. 
In \evomoe, we consider decoupling the training of experts and the gating network by involving all experts starting with a high temperature in Gumbel-Softmax and then training the gating network to be sparser and select the best expert through decaying this temperature. 
BASELayer~\cite{lewis2021base} formulates the token-expert routing as a linear assignment problem and guarantees balanced compute loads by employing numerous algorithms. 
%(Hash Layer) 
HashLayer~\cite{roller2021hash} replaces the gating network with a hash-based routing strategy (e.g., random hash, clustered hash dispersed hash).
%decouple v.s joint training
%pruning(DTS) v.s. direcly sparse 
MoEfication~\cite{zhang2021moefication} utilizes the sparse activation in feed-forward networks (FFNs) of a Transformer model and divides each large dense FFN into several experts to accelerate model inference, while EvoMoE evolves a small dense model into a large and sparse MoE model.

\paragraph{Multi-Task Learning with \moe{}} 
The multi-task learning (MTL) adopts a shared architecture to learn multiple tasks, which exploits relationships among tasks and achieve better generalization performance~\cite{DBLP:journals/corr/mtl_overview}. However, sharing parameters between unrelated tasks can potentially degrade performance.
The multi-gate MoE (i.e., MMoE) architecture is introduced as an effective way to exploit both the commonalities and differences among tasks, where each task has its own gate that adaptively controls the extent of parameter sharing~\cite{ DBLP:conf/kdd/mmoe}. 
DSelect-K~\cite{DBLP:conf/nips/dselectk} involves sparse gates (Top-K) for better parameter sharing and trains gates from dense to sparse for smoothness,
Regarding DSelect-K and the DTS gate in our \evomoe{} both propose the dense-to-sparse mechanism, they are trying to address two totally different problems, though both two pieces of work show SparseMoE is better than DenseMoE by coincidence.
For our DTS work, because DenseMoE performs well but cost for large model pretraining, we tried to find a more efficient solution (SparseMoE). While for the multi-task work, because DenseMoE performs badly for multi-task learning, they tried to find a better solution to deal with various tasks, i.e., DSelectK. Therefore, this two pieces of work have clearly different motivations.

\section{Conclusion and Future Work}
\label{sec:conclusion}
MoE models suffer from the training efficiency challenge due to the difficulty of training many experts and the gate network jointly. 
In this work, we presented an MoE training framework \evomoe that decouples the training of experts and the gate network by first spawning multiple diverse experts from one single well-trained base expert and then learning a increasingly sparse gate from a dense gate.
Our evaluations show that \evomoe can not only achieve better model quality in Transformers with given computation budget but also achieve better FLOPs-efficiency when comparing with previous works in MoE training. 
%\evomoe is a stepping stone in exploring MoE training. 
On the other hand, \evomoe opens challenges for system execution due to the computation in the early stage and the adaptive capacity of experts. 
In the future, we would like to design and implement system-level optimizations to achieve efficient training in both model quality and system execution.

\balance
\bibliography{sample}

\end{document}